%% file: main.tex
\documentclass[12pt]{article}

\usepackage{times}

\usepackage[utf8]{inputenc}
\usepackage{amsmath}
\usepackage{amssymb}
\usepackage{amsthm}
\usepackage{amsfonts}
\usepackage{bm}
\usepackage{microtype}
\usepackage{graphicx}
\usepackage[dvipsnames]{xcolor}

\usepackage{url}

\usepackage{hyperref}
\hypersetup{
    colorlinks=true,
    linkcolor={blue!50!black},
    citecolor={blue!50!black},
    urlcolor={blue!80!black}
}

\usepackage{authblk}

\usepackage[
backend=biber,
style=science,
block=ragged
]{biblatex}
\usepackage{algorithm,algorithmicx,algpseudocode}
\algblock{Input}{EndInput}
\algnotext{EndInput}
\algblock{Output}{EndOutput}
\algnotext{EndOutput}

\usepackage{afterpage}

\usepackage{lineno}
\usepackage{etoolbox}
\apptocmd{\thebibliography}{\raggedright}{}{}
\input{math_commands.tex}

\input{macros.tex}

\title{
    Building a Foundation for Data-Driven, Interpretable, and Robust Policy Design using the AI Economist
}
\author[*,1]{Alexander Trott}
\author[*,1]{Sunil Srinivasa}
\author[1]{Douwe van der Wal}
\author[2]{Sebastien Haneuse}
\author[*,${}^\dagger$,1]{Stephan Zheng}
\affil[1]{Salesforce Research}
\affil[2]{Harvard University}

\date{\today}
\begin{document}
\newrefsection
\maketitle
\input{src/v2/abstract.tex}
\let\thefootnote\relax\footnotetext{*: AT, SS, and SZ contributed equally.
${}^\dagger$: Correspondence to: \url{stephan.zheng@salesforce.com}.
AT, SS, SH, and SZ developed the theoretical framework.
DW, SS, and AT cataloged and processed data.
SS, AT, and SZ developed the simulator.
SS and AT performed experiments.
AT, SS, SH, and SZ analyzed experiments.
AT and SZ drafted the manuscript.
SS and SH commented on the manuscript.
SZ and AT conceived the project.
SZ planned and directed the project.
Source code for the economic simulation is available at \url{https://www.github.com/salesforce/ai-economist}.
More information is available at \url{https://www.einstein.ai/the-ai-economist}.
}

\input{src/v2/intro.tex}
\input{src/v2/main_body.tex}
\input{src/v2/discussion.tex}
\newpage
\setcounter{biburllcpenalty}{7000}
\setcounter{biburlucpenalty}{8000}
\printbibliography[title=References]

\input{src/v2/methods.tex}

\input{src/v2/end-notes.tex}
\input{src/v2/extended-data.tex}

\end{document}

%% file: math_commands.tex
\usepackage{amsmath,amsfonts,bm}

\def\eqref#1{equation~\ref{#1}}

\def\va{{\bm{a}}}

\def\vs{{\bm{s}}}

\DeclareMathAlphabet{\mathsfit}{\encodingdefault}{\sfdefault}{m}{sl}
\SetMathAlphabet{\mathsfit}{bold}{\encodingdefault}{\sfdefault}{bx}{n}

\def\gT{{\mathcal{T}}}

\newcommand{\E}{\mathbb{E}}

\newcommand{\eq}[1]{\begin{align}#1\end{align}}

\newcommand{\brck}[1]{\left(#1\right)}

\newcommand{\brcksq}[1]{\left[#1\right]}

%% file: macros.tex
\newcommand{\deaths}{D}

\newcommand{\infected}{I}
\newcommand{\infectionrate}{\beta}
\newcommand{\infectionratebias}{\beta^0}

\newcommand{\infectionrateslope}{\beta^\pi}

\newcommand{\mortalityrate}{\mu}

\newcommand{\population}{n}
\newcommand{\recovered}{R}
\newcommand{\recoveryrate}{\nu}

\newcommand{\stringencyindex}{\pi}
\newcommand{\stringencylevel}{\sigma}
\newcommand{\susceptible}{S}

\newcommand{\USState}{US state}
\newcommand{\vaccinated}{V}
\newcommand{\vaccinationrate}{v}
\newcommand{\dailyvaccinationrate}{\Delta V}
\newcommand{\vaccinationstartdate}{T_0}

\def\Federal{federal}

\newcommand{\statesubsidy}{\tilde{T}^{state}}

\newcommand{\unemployment}{U}

\def\costofborrowing{c}
\def\crra{\textrm{crra}}
\def\economicindex{E}

\def\healthindex{H}
\def\healthindexweight{\alpha}

\def\productivity{P}

\def\socialwelfare{F}
\def\socialwelfaremixingcoeff{\alpha}

\def\SWF{\socialwelfare}
\def\SWFweight{\alpha}

\def\util{u}

\def\mweight{\theta}
\def\pweight{\phi}

\newcommand{\policy}{\pi}

\def\ob{o}
\def\Ac{\va}
\def\ac{a}
\def\St{\vs}
\def\st{s}

\def\rew{r}

\def\df{\gamma}
\def\pol{\pi}

\def\eplen{T}

\def\trans{\gT}

%% file: src/v2/abstract.tex
\begin{abstract}
Optimizing economic and public policy is critical to address socioeconomic issues and trade-offs, e.g., improving equality, productivity, or wellness, and poses a complex mechanism design problem.
A policy designer needs to consider multiple objectives, policy levers, and behavioral responses from strategic actors who optimize for their individual objectives.
Moreover, real-world policies should be explainable and robust to simulation-to-reality gaps, e.g., due to calibration issues.
Existing approaches are often limited to a narrow set of policy levers or objectives that are hard to measure, do not yield explicit optimal policies, or do not consider strategic behavior, for example.
Hence, it remains challenging to optimize policy in real-world scenarios.
Here we show that the AI Economist framework enables effective, flexible, and interpretable policy design using two-level reinforcement learning (RL) and data-driven simulations.
We validate our framework on optimizing the stringency of \USState{} policies and Federal subsidies during a pandemic, e.g., COVID-19, using a simulation fitted to real data.
We find that log-linear policies trained using RL significantly improve social welfare, based on both public health and economic outcomes, compared to past outcomes.
Their behavior can be explained, e.g., well-performing policies respond strongly to changes in recovery and vaccination rates.
They are also robust to calibration errors, e.g., infection rates that are over or underestimated.
As of yet, real-world policymaking has not seen adoption of machine learning methods at large, including RL and AI-driven simulations.
Our results show the potential of AI to guide policy design and improve social welfare amidst the complexity of the real world.
{}
\end{abstract}

%% file: src/v2/intro.tex
\hypertarget{Introduction}{%
\section{Introduction}\label{sec:Introduction}}
Designing policy for real-world problems requires governments (\emph{social planners}) to consider multiple policy objectives and levers, and hierarchical networks of strategic actors whose objectives may not be aligned with the planner's objectives.
Moreover, real-world policies should be effective, robust to simulation-to-reality gaps, but also explainable, among others.
As such, optimizing policy is a complex problem, akin to \textit{mechanism design} \autocite{myerson1981optimal}, for which few tractable and comprehensive solutions exist.%
\footnote{
    The word ``policy'' is overloaded. First, it can refer to the decisions made by a social planner (government), as in ``public policy'' and akin to a \emph{mechanism} \autocite{myerson1981optimal}. Second, it may refer to the behavioral policy model of agents, as is common in the reinforcement learning literature. In this work, ``policy'' and ``planner policy'' refer to the social planner, while ``behavioral policy'' and ``agent policy'' refer to the behavior of agents.
}

Towards AI policy design for the real world, we extend the AI Economist framework \autocite{zheng2020ai,zheng2021ai} to learn effective, robust, and interpretable policies.
The AI Economist combines multi-agent, \emph{two-level}%
\footnote{Also referred to as ``bi-level''.}
 reinforcement learning (RL) -- e.g., training both a social planner and economic agents -- in economic simulations grounded in real-world data.
This infrastructure was previously used to design AI income tax policies that improve equality and productivity \autocite{zheng2020ai,zheng2021ai}.

We present two contributions.
First, we outline desirable properties of a policy design framework that can be used in the real world and review recent progress in machine learning and economics towards this goal.
Second, we apply this framework to a proof-of-concept use case: designing \USState{} and \Federal{} policy to respond to the COVID-19 pandemic, evaluated within an integrated pandemic-economic simulation based on real data.

The AI Economist framework fills an important gap in the space of policy design.
Existing analytical approaches are limited to a narrow set of policy objectives or do not yield explicit solutions in the sequential setting.
Empirical methods lack counterfactual data to estimate behavioral responses (the \emph{Lucas critique}).
Moreover, existing simulation-based approaches often do not consider both strategic planners and agents, or interactions between actors \autocite{fernandez-villaverde_econometrics_2009,benzell_simulating_2017,Hill2021}.
Such complex settings include income taxation (tax policy changes the post-tax utility that economic agents experience) and pandemics (Federal subsidies can lighten the economic burden on citizens and may incentivize \USState{}s to employ more stringent public health policies).

As a case study, we show how our framework can optimize public health and economic policies for both public health (e.g., reducing COVID-19 deaths) and the economy  (e.g., maintaining productivity) during a pandemic.
Using real data, we simulate COVID-19, vaccinations, and unemployment in the United States.
We model the stringency level of 51 \USState{}-level policies%
\footnote{
    We explicitly mention ``US'' in ``\USState{}'' to disambiguate it from ``state'' as used in the RL context.
}, including all 50 \USState{}s and Washington D.C.,
and \Federal{} subsidies in the form of direct payments to citizens.
More stringent \USState{} policy (e.g., closing down businesses more) may temper the spread of the pandemic \autocite{flaxman_estimating_2020,acemoglu_optimal_2020}, but may lead to lower productivity owing to higher unemployment.
As a modeling assumption, the policy objective of \USState{}s depends on the level of \Federal{} subsidies: higher \Federal{} subsidies may incentivize \USState{}s to accept higher unemployment and be more stringent to reduce deaths.
However, this may incur a higher national cost, e.g., through borrowing money.
Hence, this presents a hard two-level optimization problem.
Although the underlying public health and economic mechanics are complex, these correlations and trade-offs are salient in real data.

We find that log-linear RL policies achieve significantly higher social welfare than baseline policies executed in simulation.
These policies perform well across a wide range of relative weightings of public health and the economy, and demonstrate how the social planner can align \USState{} incentives with national social welfare through subsidies.
The policies are explainable: well-performing policies respond strongly to changes in recovery and vaccination rates, for example.
Moreover, these results are robust to a range of calibration errors, e.g., infection rates that are over or underestimated.

Our work is descriptive, not normative.
We show outcomes for a class of stylized social welfare objectives which follows standard economic modeling.
However, we acknowledge that the objectives of real-world actors typically involves many features.
Taken together, our results motivate building representative and fine-grained simulations and developing machine learning algorithms to learn AI-driven policies that can \emph{recommend} policy and improve social welfare in the real world.

%% file: src/v2/main_body.tex
\begin{figure}[t!]
\centering
    \includegraphics[width=\linewidth]{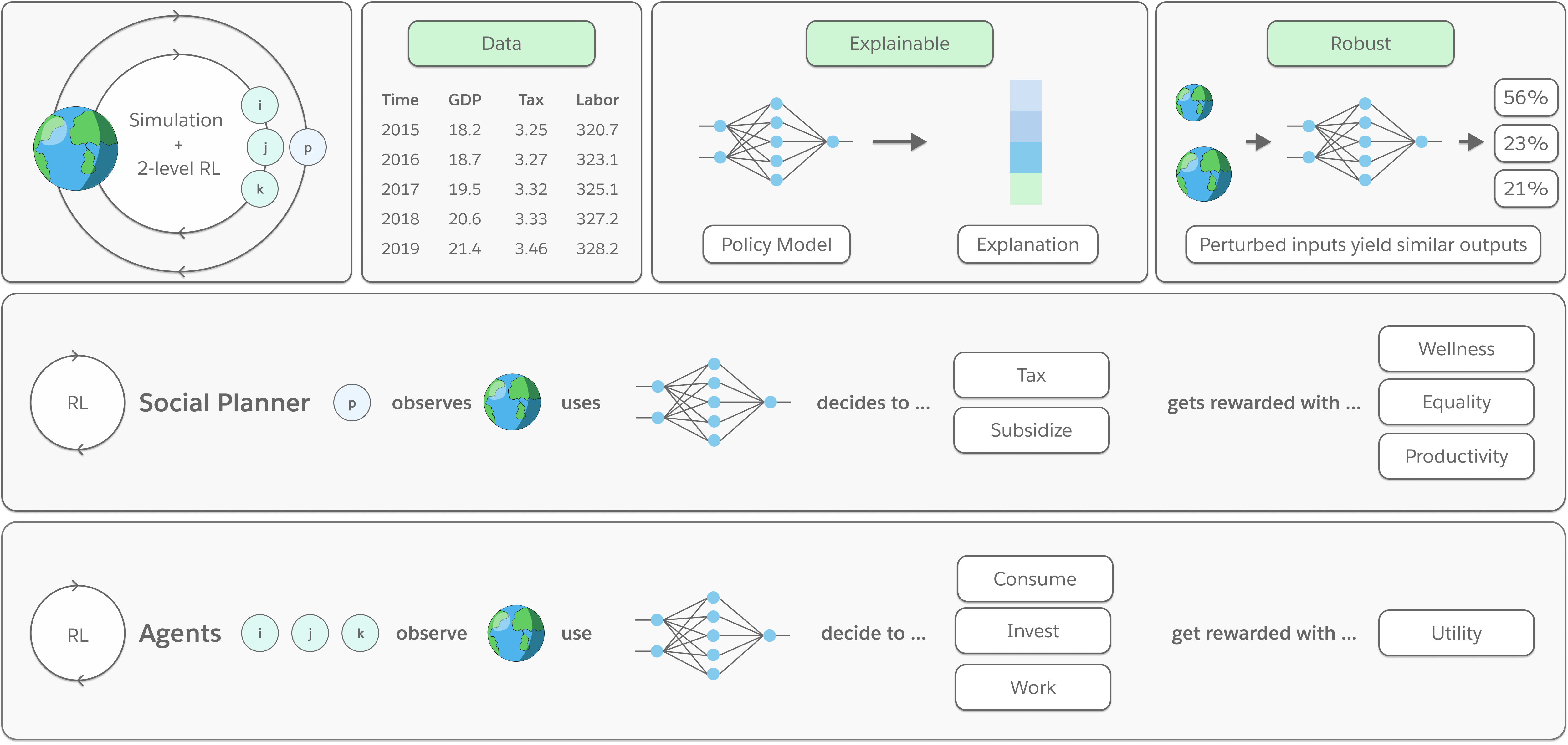}
    \caption{
        \textbf{Extending the AI Economist framework for policy design.}
        The core of the AI Economist framework is an economic simulation featuring a social planner and economic agents, together called actors.
        All actors optimize their (behavioral) policy model using reinforcement learning; repeatedly observing, deciding, and updating their model using reward feedback.
        As agents may strategically respond to the planner's policy, this poses a two-level (or \emph{bi-level}) optimization problem.
        We extend the framework studied in \autocite{zheng2020ai,zheng2021ai} by adding key desirable properties for AI policy design for the real world.
        These include using simulations fitted to real data, using explainable policy models, and ensuring robust performance under meaningful simulation parameter perturbations, see Section \ref{sec:AI-driven Policy Design for the Real World}.
        We use a case study on pandemic response policy to demonstrate these features, see Section \ref{sec:Case Study: Optimizing Public Health and the Economy during Pandemics}.
        Further extensions and aspects of our framework are discussed in Section \ref{sec:Discussion}.
    }
    \label{fig:general-framework}
\end{figure}
\hypertarget{AI-driven Policy Design for the Real World}{%
\section{AI-driven Policy Design for the Real World}\label{sec:AI-driven Policy Design for the Real World}}
AI methods can help solve complex policy design problems, as illustrated in our case study on optimizing pandemic response policies in Section \ref{sec:Case Study: Optimizing Public Health and the Economy during Pandemics}.
Our AI policy design framework consists of two pillars: \emph{simulations}, grounded in real-world data and featuring AI agents, and \emph{two-level RL},
as outlined in Figure \ref{fig:general-framework}.
In this Section, we describe how machine learning can be applied to economic policy design, aligning concepts from both fields.
We then describe several features of the AI Economist framework that are desirable for designing policy, and highlight open challenges.
We also outline how to extend our framework to increase realism in the future in the Discussion.

\subsection{A Learning Perspective on Economic Policy Design}
In this work, we consider a dynamic environment with two \emph{actor} types: a social planner (denoted by $p$) and $N$ economic agents. We index actors with $i = 1, \ldots, N, p$.
For example, the social planner represents a government, while the economic agents represent individuals.
In our Case Study, the social planner represents the federal government, while the economic agents are the \USState{}s.

Formally, the environment can be represented as a Markov Game \autocite{littman1994markov}.
Episodes iterate for $\eplen > 1$ time steps; at each time $t$, each actor $i$ observes a part of the world state $\St_t$, can perform actions $\Ac_{i,t}$, and receives a scalar reward $\rew_{i,t}$.
The environment then transitions to the next state $\St_t$ using environment dynamics $\trans\brck{\St_{t+1} | \St_t, \Ac_t}$.
Assuming rationality, each actor optimizes its behavioral policy $\pol{}_i\brck{\Ac_{i,t}|\ob_{i,t}}$ to maximize its total (discounted) future reward $\sum_{t}\df_i^t\rew_{i,t}$. Here, $\ob_{i,t}$ is the observation received by $i$ at time $t$, and $0 < \df_i \leq 1$ is a discount factor.
A small (or large) discount factor, $\df_i\approx 0$ (or $\df_i\approx 1$), emphasizes long-term (or short-term) rewards.

Following standard economic modeling, each agent experiences a \emph{utility} $\util_{i,t}$; its reward is its marginal utility gain $\rew_{i,t} = u_{i,t} - u_{i, t-1}$.
In effect, an agent optimizes its total (discounted) utility over the course of an episode.%
\footnote{
In dynamic economic models, utility is often modeled as a function of the instantaneous consumption at a single time step.
In that case, each step represents a period with meaningful total consumption, e.g., a month or year.
In our work, each time step represents a day, a more fine-grained temporal resolution, and episodes last for about a year.
As such, in our setting, it is more natural to define rewards such that we accumulate utility over an entire episode.
Here, each episode can be thought of as a single time step at a coarser temporal resolution.
}
The social planner aims to optimize \emph{social welfare} $\SWF\brck{\St_{0}, \ldots, \St_{\eplen}; \bm{\SWFweight}}$, parameterized by $\bm{\SWFweight}$, which, informally speaking, represents how well off society is.
Accordingly, the planner's reward at time $t$ is the marginal gain in social welfare.
A common social welfare choice is the utilitarian objective, in which the planner optimizes $\sum_{t}\df_p^t \sum_{i}\util_{i,t}$, using a discount factor $0 < \df_p \leq 1$.

In an economic context, rewards represent incentives and may incentivize agents to cooperate or compete.
As such, the combination of planner and agent reward functions $\brck{\rew_1, \ldots, \rew_\population, \rew_p}$ can present significant learning challenges and yield complex behaviors during learning and at equilibrium.
For instance, the social planner may attempt to align \emph{a-priori} misaligned incentives between the planner and agents, as studied in principal-agent problems \autocite{grossman1992analysis}.

\paragraph{Two-Level Learning.}
A special feature of the policy design problem is that it presents a \emph{two-level learning problem}: the actions of an actor can explicitly affect the reward that other actors optimize for.
For example, a planner that changes income tax rates changes the post-tax income that agents receive, which can change (the shape of) their utility \autocite{zheng2020ai,zheng2021ai}.
As another example, in this work we consider how planner subsidies (\Federal{} level) can affect the social welfare experienced by agents (\USState{} level).
As such, two-level learning problems and generalizations thereof appear naturally in many economic and machine learning settings \autocite{goodfellow2014generative}.
However, they also present a hard optimization challenge and can lead to highly unstable learning behavior.
The key reason is that as one actor changes its policy, the shape of the entire reward function for other actors may change significantly, and cause actors to change their behavior.
Hence, our policy design setting is distinct from games in which deep RL has reached (super)-human skill, e.g., Go \autocite{silver2017mastering} and Starcraft \autocite{vinyals2019grandmaster}, but that have a fixed reward function throughout.

\paragraph{Mechanism Design.}
The planner's policy design problem is an instance of \emph{mechanism design} \autocite{myerson1981optimal} with strategic agents.
A mechanism constitutes a set of rewards and environment dynamics \autocite{vickrey1961counterspeculation}, which is akin to the planner's policy.
The mechanism designer's goal is to find a mechanism that certain desired properties, e.g., it incentivizes agents to reveal their true preferences to the planner.
However, few tractable analytical solutions to the general mechanism design problem exist, e.g., a full analytical understanding is lacking for auctions with even two items and multiple bidders \autocite{daskalakis2015multi}.
As such, our work presents an RL-based solution to find optimal mechanisms.

\subsection{Features of AI-driven Policy Design}
\label{sec:frameowrk-features}
The use of AI methods, including RL, has benefits for policy design, but also has idiosyncratic features that present practical challenges and remain open areas of research.
We now present how these benefits are manifest in the AI Economist framework, how it addresses the challenges of using AI for policy design, and how it compares to existing policy design methods.

\paragraph{Data-Driven Economic Simulations.}
Existing computational approaches to economic modeling include general equilibrium models \autocite{cox1985intertemporal,smets2003estimated} and inter-generational inter-regional dynamics \autocite{benzell_simulating_2017} that consider macro-economic, aggregate phenomena.
However, these approaches typically do not model the behaviors of and interactions between individuals, while their structural estimation and calibration to real data remains challenging.

On the other hand, agent-based models \autocite{railsback2019agent,macal2005tutorial}, including agent computational economics \autocite{tesfatsion2006agent}, study economies as dynamic systems of individual agents and the resulting emergent phenomena.
However, previous work in these areas largely lack rich behavioral models for agents that implement strategic behaviors, for example, using multi-agent reinforcement learning.

Our learning-based approach builds on these ideas by modeling strategic behavior at all levels, e.g., training social planner and economic agents who optimize for their individual or social objectives.
As such, our framework enables a more richer descriptions of collective behavior and emergent phenomena.

\paragraph{Using Real Data.}
Similar to standard economic modeling, our learning approach uses economic simulations that are fitted to real data.
Beyond the exposition in this work, we outline more ways to learn more realistic economic models using micro-level and macro-level data.
First, micro-level data could be used to learn models that mimic human behavior, learning and adaptation, and utility functions.
Second, macro-level data could be used to fit the aggregate (emergent) behaviors of collections of strategic agents.
See the Discussion for more details.

\paragraph{Finding Equilibrium Solutions.}
In a game-theoretic sense, a key objective of policy design is to find equilibrium solutions in which the behavior of all actors is optimal.
Optimality is not uniquely defined in the multi-agent setting; many definitions of equilibria exist.
In this work, we assume the planner's policy should maximize social welfare and agents should maximize their utility given the planner's policy.
Such two-level structures (planner and agents) are akin to Stackelberg games \autocite{von2010market}.
As both the agents and planner are strategic, finding equilibrium solutions presents a challenging learning problem.

The AI Economist uses RL to solve the two-level problem, learning both the optimal planner's policy and agent behavioral policies using RL.
Given the complexity of sequential two-level optimization, few theoretical guarantees for finding exact equilibria exist.
Moreover, it is hard to enumerate or classify all equilibria for general-sum, two-level problems.
However, we find that two-level RL can find well-performing policies that are both robust and explainable.
Specifically, several intuitive two-level RL strategies have been shown to be effective, including the use of curricula and training actors separately or jointly in multiple phases \autocite{zheng2020ai,zheng2021ai}.

\paragraph{Flexible Policy Objectives and Levers.}
In the real world, economic policy needs to be optimized for a wide spectrum of policy objectives, such as productivity, equality, or wellness.
However, economic approaches often optimize policy for total utility and other stylized objectives and constraints \autocite{mascolell_microeconomic_1995}.
Such objectives often enjoy certain theoretical guarantees or analytical properties, but might be unrealistic and can be hard to measure in the real world.
For instance, it is hard to measure utility, while human agents might not act optimally or display behavior that optimizes a stylistic utility function.

Similarly, it can be challenging to optimize policy across multiple levers, such as taxes, subsidies, or business closures.

A key benefit of using RL is that we can use any quantitative objective, including those that are not analytical or differentiable.
For example, the AI Economist optimizes both public health and economic objectives in the Case Study.
This makes our framework highly \emph{flexible} and easier to align with real-world use cases.

\paragraph{Robustness.}
A critical aspect of economic modeling is uncertainty due to a lack of data, noisy data, structural misspecification, lack of model capacity, or other issues.
Specifically in the machine learning context, policies learned by RL may be not transfer well to the real world if they are not robust to errors in the underlying economic simulations and gaps between the simulation and the real world.

In this work, we demonstrate empirically that two-level RL can learn log-linear policies that are robust under perturbations of simulation parameters, see the Case Study for details.
However, it is still hard to theoretically guarantee (levels of) robustness in the general setting, and especially when using complex model classes such as deep neural networks \autocite{shi2019neural}.
Several machine learning techniques could be used to improve the robustness of policy models,
including data augmentation \autocite{zheng2016improving}, domain randomization \autocite{tobin2017domain}, adversarial methods \autocite{pinto2017robust}, and reward function perturbation \autocite{zhao2021ermas}.
Furthermore, when using linear policies and linear simulation dynamics, control theoretic guarantees may exist \autocite{doyle1988state}.
In all, improving robustness remains a key research area.

\paragraph{Explainability.}
Machine learning models have shown to be highly effective for many prediction and behavioral modeling problems.
In particular, deep neural networks can learn rich strategic behaviors in economic simulations \autocite{zheng2020ai,zheng2021ai}.
However, to build trust, learned economic policies should also be \emph{explainable} or \emph{interpretable}.
Unfortunately, there is no general consensus definition of explainability for high-dimensional machine learning models.
Rather, explanations are often domain-specific \autocite{doshi2017towards} and their acceptability is prone to subjectivity.

A form of explainability is feature attribution.
For example, the weights of linear models show how predictions, e.g., policy decisions, depend on (a change in) the input features.
Using this rationale, we show in our Case Study that log-linear policies that achieve high social welfare respond strongly to (rises in) infection rates.

By extension, local function approximations may provide \emph{post-hoc} explanations of a complex model's behavior on a subset of the data \autocite{ribeiro2016should}.

However, one must guard against over-interpreting explanations, especially in high-dimensional state or action spaces, and for complex policy models, such as deep neural networks.
For instance, ``plausible'' explanations may not be robust \autocite{slack2020fooling} or generalize to all data or unseen environments.

\paragraph{Simplicity and Policy Constraints.}
A heuristic quality of real-world policy is that it should be ``simple'' enough to be acceptable. Furthermore, ``simple'' policies may be more explainable or robust.
For example, it could be desirable that real-world policy should not change too often or too dramatically, e.g., doubling tax rates.
This parallels common economic modeling choices, such as consumption smoothing \autocite{morduch1995income} and sticky prices \autocite{mankiw2002sticky}.
A benefit of our RL approach is that it is relatively straightforward to impose constraints on policy, e.g., through regularization or manually specified action masking.
We find empirically that gradient-based optimization techniques can effectively find optimal policies under such constraints.
Moreover, more sophisticated techniques for constrained RL could be used \autocite{achiam2017constrained}.
In contrast, it can be hard for analytical policy design frameworks to include constraints that are not differentiable or analytical.

\paragraph{Challenges of RL for Policy Design.}
Although RL is a flexible framework that can find well-performing policies, there are several learning challenges that are topics of active research.
One salient challenge is that \emph{model-free} RL, as used in our work, can be inefficient: it typically requires many learning samples acquired through interaction with the simulation.
Here, ``model-free RL'' refers to RL methods that do not use or learn structural models of the simulation environment \autocite{sutton2018reinforcement}.
Also, multi-agent RL can be challenging because any RL agent experiences and learns in a non-stationary environment when other RL agents are also learning and changing their behavior \autocite{bucsoniu2010multi}.
Despite these challenges, we find empirically that RL policies can be trained to perform well in our Case Study within a reasonable amount of time, e.g., 1-2 days, using the learning strategies detailed hereafter.

\begin{figure}[t!]
\centering
    \includegraphics[width=\linewidth]{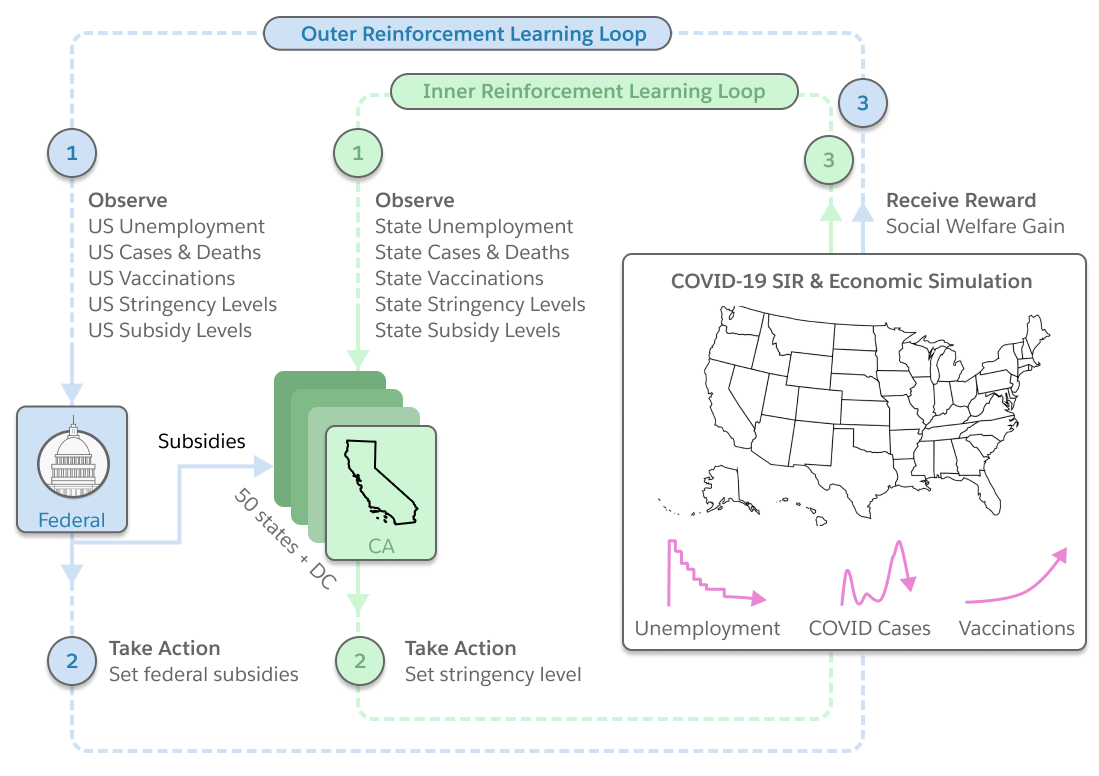}
    \caption{
        \textbf{Policy design using a simulation of the COVID-19 pandemic and the economy, and two-level reinforcement learning.}
        We train policy models for a social planner (\Federal{} government) and agents (\USState{}s).
        Each actor takes actions using its policy model, based on an observation of an integrated pandemic-economic simulation of the United States.
        \USState{} agents (\Federal{} planner) are rewarded by increases in state-level (\Federal{}-level) social welfare.
        As a modeling choice, \Federal{} subsidies can offset the economic impact for \USState{}s, which may incentivize \USState{}s to respond more stringently to the pandemic.
        As such, this poses a two-level RL problem in which \USState{}s optimize their policy, partially in response to \Federal{} policy.
    }
    \label{fig:covid-policy-design-framework}
\end{figure}
\hypertarget{Case Study: Optimizing Public Health and the Economy during Pandemics}{%
\section{Case Study: Optimizing Public Health and the Economy during Pandemics}\label{sec:Case Study: Optimizing Public Health and the Economy during Pandemics}}

We now demonstrate how to apply the AI Economist framework to design and evaluate economic and public health policy during the COVID-19 pandemic in the United States.
We model policy at the level of \USState{}s, which control the stringency of public health measures, as well as the \Federal{} government, which controls direct payment subsidies.
Each actor optimizes its own policy for its own definition of social welfare, taken here as a weighted combination of health and economic outcomes (related to deaths and GDP, respectively) within its jurisdiction.
This trade-off between public health and economic outcomes is salient in observable data collected during the pandemic, although the underlying public health and economic mechanics are complex.%
\footnote{
    This trade-off also has been a specific focus of policymakers and policy measures in public discourse.
}
The following sections illustrate the features of our proposed framework via this case study.
We present an overview of how we design the policy simulation and calibrate it with real-world data.
In addition, we analyze the AI policies and outcomes corresponding to a range of possible policy objectives.
Finally, we examine the properties of interpretability and robustness by examining the learned policy weights and analyzing model sensitivity, respectively.

\paragraph{Related Work.}
Relatively few works have studied optimal policy design in the intersection of public health and the economy.
To the best of our knowledge, no other work has studied the interaction between unemployment, COVID-19, and policy interventions.
Existing public health research has focused on many aspects of COVID-19, including
augmenting epidemiological models for COVID-19 \autocite{li_global_1995,zou_epidemic_2020},
and analyzing contact-tracing in agent-based simulations \autocite{alsdurf_covi_2020},
and evaluating the efficacy of various public health policy interventions \autocite{kapteyn_tracking_2020,flaxman_estimating_2020,chang_mobility_2021}.
An analytical approach to policy design showed that the elasticity of the fatality rate to the number of infected is a key determinant of an optimal response policy \autocite{RePEc:nbr:nberwo:26981,acemoglu_optimal_2020}.
Moreover, statewide stay-at-home orders had the strongest causal impact on reducing social interactions \autocite{Abouk2020.04.07.20057356}.

In the economics literature,
the effect of pandemic response policies has been studied, such as the relationship between unemployment insurance and food insecurity \autocite{raifman_association_2021},
while various sources have tracked government expenditure during the pandemic \autocite{irseconomicimpactpayments,usaspendinggov,covidmoneytrackerorg}.
Difference-in-difference analysis of location trace data finds that imposing lockdowns leads to lower overall costs to the economy than staying open \autocite{RePEc:oxf:wpaper:910}, under a modified SIR model.
Early US data has also shown the unequal distribution and unemployment effects of remote work across industries \autocite{RePEc:nbr:nberwo:27344}.

\hypertarget{Simulation Design and Calibration}{%
\subsection{Simulation Design and Calibration}\label{sec:Simulation Design and Calibration}}
For our case study, we built a simulation model that captures the impact of policy choices on the overall economic output and spread of the disease.
Our modeling choices are driven by what can be reasonably calibrated from publicly available health and economic data from 2020 and 2021.
Future simulations could be expanded with the availability of more fine-grained data.
This section provides an overview of the simulation design and calibration; concrete details are provided in the Methods (Sections~\ref{sec:methods-notation-and-definitions}-\ref{sec:methods-calibrating-policy-priorities-from-data}).

\paragraph{Policy Levers}
The simulation models the consequences of 51 state-level policies, for all 50 \USState{}s and the District of Columbia, and 1 \Federal{} policy.
Each state-level policy sets a \emph{stringency level} (between $0\%$ and $100\%$), which summarizes the level of restrictions imposed (e.g. on indoor dining) in order to curb the spread of COVID-19.
The stringency level at a given time reflects the number and degree of active restrictions.
This definition follows the Oxford Government Response Tracker \autocite{hale2021global}.
The \Federal{} policy periodically sets the daily \emph{per capita subsidy level}, or direct payment.
Federal subsidies take the form of direct payments to individuals, varying from $\$0$ to $\$55$ per day per person, in 20 increments.

\paragraph{Epidemiology Model.}
We use an augmented SIR-model \autocite{kermack_contribution_1927} to model the evolution of the pandemic, including the effect of vaccines.
We only model the outbreak of a single variant of COVID-19.
The standard SIR model emulates how susceptible individuals can become infected and then recover.
By convention, recoveries include deaths.
As a simplifying assumption, only susceptible individuals are vaccinated, recovered individuals cannot get reinfected, and vaccinated individuals gain full immunity to COVID-19, directly moving from susceptible to recovered\footnote{This is motivated by empirical results that vaccines commonly used in the US are more than 95\% effective at preventing serious illness due to and spreading of COVID-19.}.
Within each \USState{}, the infection rate (susceptible-to-infected) is modeled as a linear function of the state's stringency level.
This reflects the intuition that imposing stringent public health measures can temper infection rates.

\paragraph{Economic Model.}
For each \USState{}, daily economic output is modeled as the sum of incoming \Federal{} subsidies plus the net production of actively employed individuals.
At the \Federal{} level, this output is taken as the sum of the State-level outputs, minus borrowing costs used to fund the outgoing subsidies.
We model unemployment using a time-series model that predicts a state's unemployment rate based on its history of stringency level increases and decreases\footnote{We also account for deaths and active infections when modeling unemployment.}.
The daily productivity per employed individual is calibrated such that yearly GDP at pre-pandemic employment levels is equal to that of the US in 2019.

\paragraph{Datasets and Calibration.}

Data for the daily stringency policies are provided by the Oxford COVID-19 Government Policy Tracker~\autocite{hale2021global}.
The date and amount of each set of direct payments issued through \Federal{} policy are taken from information provided by the COVID Money Tracker project~\autocite{CovidMoneyTracker}.
We use the daily cumulative COVID-19 death data provided by the COVID-19 Data Repository at Johns Hopkins University~\autocite{Dong2020} to estimate daily transmission rates.
Daily unemployment rates are estimated based on the monthly unemployment rates reported by the Bureau of Labor Statistics~\autocite{BureauofLaborStatistics2021}.
We fit the disease (unemployment) model to predict each State's daily transmission (unemployment) rate given its past stringency levels.
We allow State-specific parameters for both models, but during fitting we regularize the State-to-State variability to model common trends across states and to prevent overfitting to noisy data.

We calibrated the simulation on data from 2020 only.
We fit simulation models on data through November 2020 and use December 2020 for validation.
As such, alignment between simulated outcomes during 2021 and real-world data from 2021 reflects the ability of the simulation to \textit{forecast} policy consequences beyond the timeframe used for calibration.
See Section~\ref{sec:methods-data-and-model-calibration} in the Methods for additional details regarding datasets and calibration.

\paragraph{Metrics and Objectives.}
Each actor (i.e. a \USState{} or the \Federal{} government) sets its policy so as to optimize its social welfare objective, which can be defined differently for different actors.
For each actor $i$, we assume that its social welfare objective $\socialwelfare_i$ is defined as a weighted sum of a a public health index $\healthindex_i$ and an economic index $\economicindex_i$:
\eq{
    \SWF_i(\healthindexweight_i) &= \healthindexweight_i \healthindex_i + (1 - \healthindexweight_i) \economicindex_i.
}
Here $i\in \{ 1, \ldots, 51, p \}$ indexes the and 51 \USState{}- and \Federal{}-level actors, and $\healthindexweight_i \in [0, 1]$ parameterizes the priority actor $i$ gives to health versus the economy.
With this notation, we consider the \Federal{} actor as the planner.

Each index summarizes the health or economic outcomes experienced during the simulation.
More precisely, $\healthindex_i$ (or $\economicindex_i$) is measured as the average \textit{daily} health (or economic) index for actor $i$.
The marginal health index $\Delta\healthindex_{i,t}$ at time $t$ (each time-step represents a day) measures the number of new COVID-19 deaths in the jurisdiction of actor $i$.
The marginal economic index $\Delta\economicindex_{i,t}$ is a concave function of total economic output (described above) at time $t$ in jurisdiction $i$.
\eq{
    \Delta\economicindex_{i,t} = \crra \left( \frac{\productivity_{i, t}}{\productivity_i^0} \right)
}
Here, $\productivity_{i, t}$ denotes the total economic output, which includes any incoming subsidies, and $\productivity_{i}^0$ is average pre-pandemic output.
The \Federal{} government uses
\eq{
    \productivity_{p,t} \triangleq \sum_{i=1}^N \productivity_{i, t} - \costofborrowing \cdot \statesubsidy_{i, t},
    \qquad
    \productivity_{p}^0 \triangleq \sum_{i=1}^N \productivity_{i}^0.
}
so that \Federal{}-level economic output accounts for the borrowing cost of funding outgoing subsidies $\statesubsidy$.

Using the CRRA function (see Section~\ref{sec:methods-social-metrics-and-indices}) to define $\Delta\economicindex_{i,t}$ as a \textit{concave} function of total economic output is an important modeling choice (albeit a common one in economic modeling).
In particular, it captures the intuition that a decrease in economic output is felt more severely when that output is already low.
Consequently, \Federal{} subsidies, which raise the level economic output for \USState{}s, can ``soften the blow'' of additional unemployment and thereby indirectly incentivize additional stringency by mitigating the trade-off that States face between $\healthindex_i$ and $\economicindex_i$.
This choice therefore imparts a \textit{two-level} problem structure between the \USState{}s and the \Federal{} government, further motivating the application of our policy design framework.

To simplify notation and analysis, we normalize indices so that each minimum-stringency policy yields $\healthindex_i=0$ and $\economicindex_i=1$ and so that each maximum-stringency policy yields $\healthindex_i=1$ and $\economicindex_i=0$.
As a result, higher index values denote more preferable outcomes.

Finally, we calibrate each $\healthindexweight_{i}$ by setting it to the value $\hat{\healthindexweight}_i$ that maximizes social welfare $\SWF_i(\hat{\healthindexweight}_i)$ under the real-world stringency policies (see Section~\ref{sec:methods-calibrating-policy-priorities-from-data} in the Methods for details).
Calibrating $\hat{\healthindexweight}_i$ in this way facilitates comparison against outcomes simulated using real-world policies.
However, our framework allows us to explore outcomes over a wide space of alternative $\healthindexweight$ configurations, which we demonstrate below.

\hypertarget{Results}{%
\subsection{Results}\label{sec:Results}}

We compare simulated outcomes using real-world and AI policies, trained using the AI Economist framework with the $\hat{\healthindexweight}_i$ that maximize social welfare under the real-world policies.
We use log-linear policies:
\eq{\label{eq:stringency-policy}
    \pi(\stringencylevel_{i,t} = j | \ob_{i,t}) = \frac{1}{Z_{i,t}} \exp \left( \sum_{k}\ob_{ik,t}W_{kj} + b_{ij} \right).
}
Here, $\pi(\stringencylevel_{i,t} = j | \ob_{i,t})$ denotes the (conditional) probability that \USState{} $i$ will select stringency level $\stringencylevel_{i,t} = j$ at time $t$, given its observations $\ob_{i,t}$.
We discretize the stringency level $\stringencylevel$ into 10 levels, such that $j \in \{1, \ldots, 10 \}$ indexes each of the possible stringency levels.
The normalization $Z_{i,t}$ ensures that $\pi$ is a proper probability distribution for all $i$ and $t$: $\sum_j \pi(\stringencylevel_{i,t} = j | \ob_{i,t}) = 1$.
The terms $W$ and $b$ represent the learnable parameters that are optimized when training the AI policies.
$W_{kj}$ is the weight matrix, with $j$ indexing stringency levels and $k$ indexing input features.
$b_{ij}$ is the bias that \USState{} $i$ has towards stringency level $j$.
The weight matrix $W$ is shared across \USState{}s, whereas the bias terms $b$ are specific to each state.
A similar model is used for the \Federal{} policy.
This policy model choice emphasizes explainability and simplicity, which we explore in more detail hereafter.

Throughout training and analysis, we initialize each simulation episode such that $t=0$ corresponds to March 22, 2020.
During training, episodes run for $T=540$ timesteps; however, at the time of this analysis, real-world data were only available through the end of April 30, 2021 ($t = 404$), so we treat this date as the end of our analysis window.

\begin{figure}[t!]
\centering
    \includegraphics[width=\linewidth]{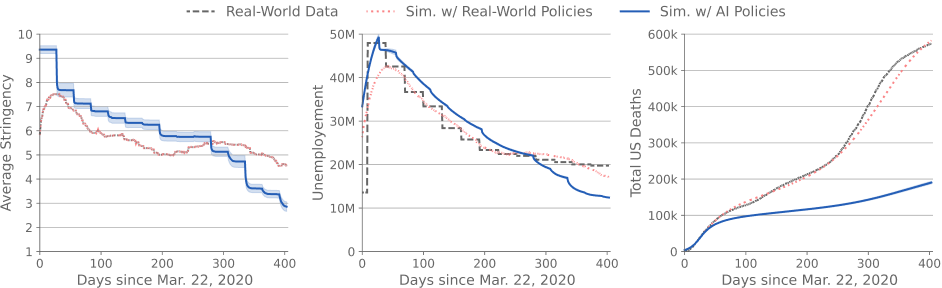}
    \caption{
        \textbf{Real-world vs simulation outcome.}
        Left: average \USState{} policy stringency level.
        Middle: US unemployment rate.
        Right: total US deaths.
        AI policies impose more stringent policy at the start of the pandemic before tapering off more quickly.
        This temporarily yields higher unemployment, but results in fewer total deaths.
        The simulation fits the data well: the simulation fits the real data well when executed with the real-world policy.
    }
    \label{fig:basic-comparison}
\end{figure}

\begin{figure}[t!]
\centering
    \includegraphics[width=\linewidth]{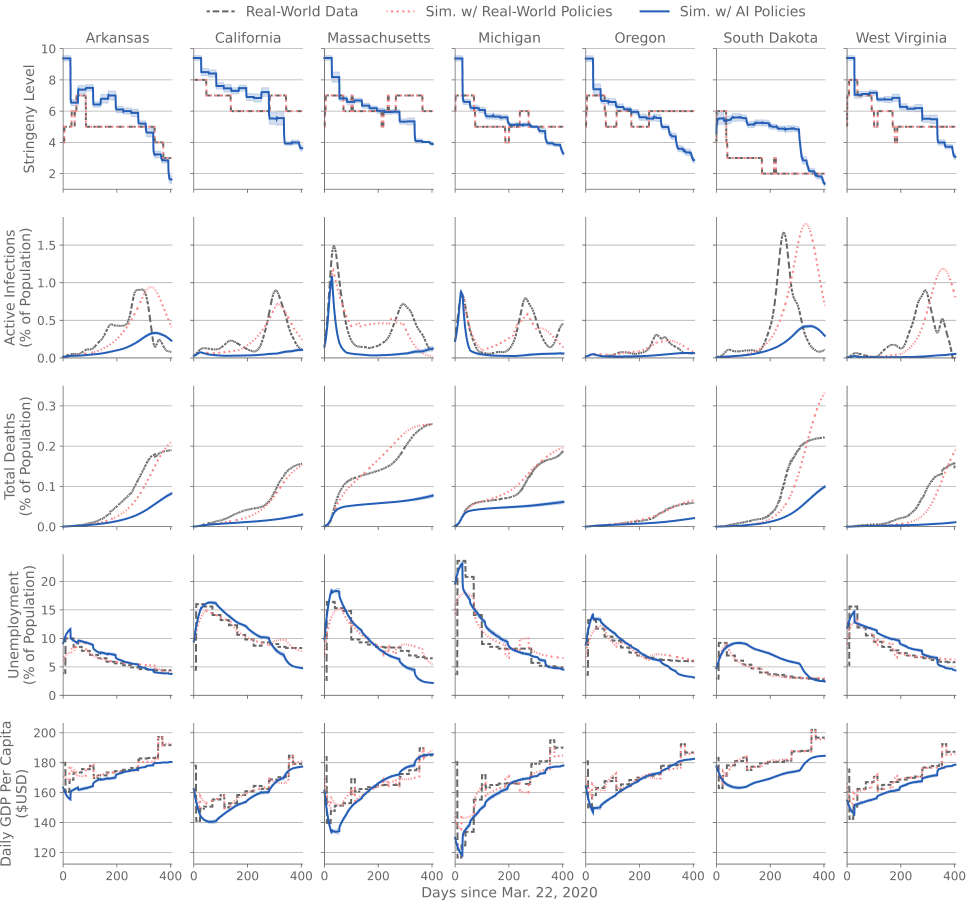}
    \caption{
        \textbf{Real-world vs simulation outcome at the state level.}
        In order, from top to bottom: Stringency level; active COVID-19 case load; cumulative COVID-19 deaths; unemployment; and daily economic output.
        Infections, deaths, and unemployment numbers are expressed as percentages of the state population; economic output is similarly normalized to reflect \textit{per capita} daily GDP.
        Note: For the ``Real-World'' data (gray), infection numbers are estimated based on available data on COVID-19 deaths (see Methods) and economic output is estimated from our model given real-world data as inputs.
    }
    \label{fig:state-comparison}
\end{figure}

As shown in Figure~\ref{fig:basic-comparison}, simulated unemployment and COVID-19 deaths under real-world policies (dashed red lines) approximate the real-world trends well (dashed gray lines).
Compared to real-world policies, AI policies (blue lines) impose comparatively higher stringency at the start of the outbreak but reduce stringency more rapidly.
Similarly, AI policies result in more unemployment early on but recover towards pre-pandemic levels more quickly.
Overall, however, unemployment under AI policies is higher on average during the analysis window.
Moreover, AI policies result in considerably fewer COVID-19 deaths in this simulation.
Figure~\ref{fig:state-comparison} illustrates these trends for several \USState{}s. For a full view including all \USState{}s, see the Extended Data.

\subsection{Improved Social Welfare}

\begin{figure}[t!]
    \centering
    \includegraphics[width=\linewidth]{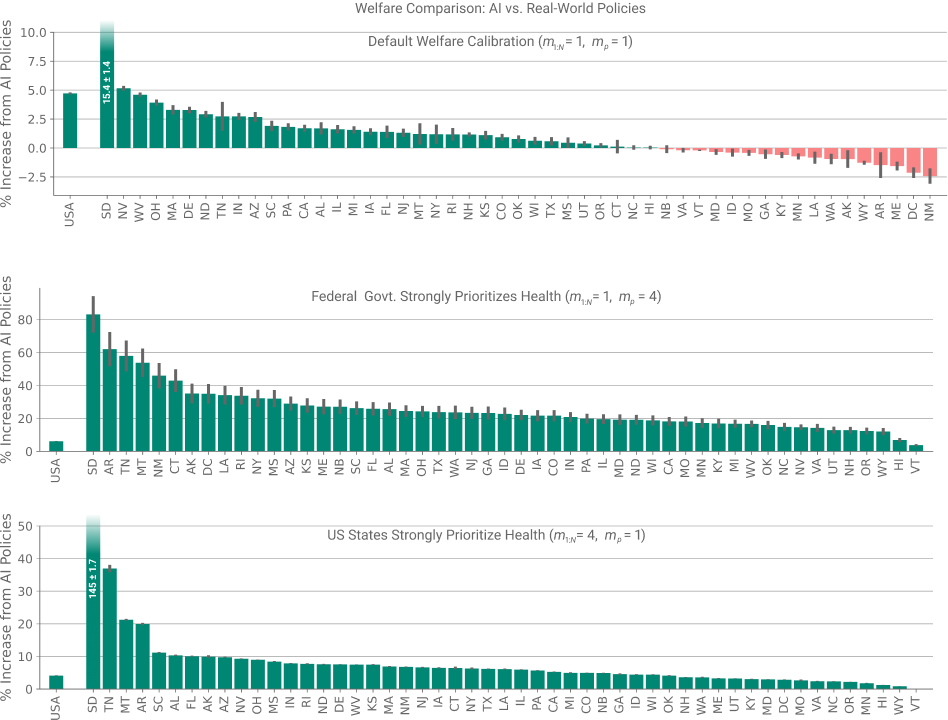}
    \caption{
        \textbf{AI Policies Achieve Higher Social Welfare in Simulation.}
        AI policies achieve higher social welfare at the \Federal{} level and for 33 out of 51 \USState{}s when compared to real-world policies.
        Top: welfare comparison when using $\hat{\healthindexweight}$ values given by the default calibration.
        Middle: welfare comparison with highest tested $\healthindexweight_p$ and default $\hat{\healthindexweight}_{1:N}$.
        Bottom: welfare comparison with highest tested $\healthindexweight_{1:N}$ and default $\hat{\healthindexweight}_p$.
        In each plot, welfare is calculated with the $\healthindexweight$ values used to train AI policies.
        Bar heights and error bars denote the mean welfare improvement (as a percentage of welfare under real-world policies) and STE, respectively, across the 10 random seeds used to train AI policies, for each actor in the simulation.
    }
    \label{fig:improved-social-welfare}
\end{figure}

Figure~\ref{fig:improved-social-welfare} shows the percentage change in welfare, for each actor, from AI policies versus from real-world policies in the simulation.
The top subplot shows results when social welfare is defined using the $\hat{\healthindexweight}$ values obtained during calibration (see above).

Federal-level welfare is 4.7\% higher under AI policies.
We identify two features underlying this improvement.
First, the AI stringency policies achieve a more favorable balance between unemployment and COVID-19 deaths.
Second, the AI subsidy policy provides very little direct payments in order to achieve this balance.

In comparison, the real-world policies include a total of \$630B in subsidies (versus just \$35B on average for AI policies).
\USState{}-level actors \textit{gain} welfare via subsidies.
As a result, the welfare benefits from AI policies are less pronounced at the level of \USState{}s.
Nevertheless, we observe that welfare improves for 33 of the 51 state-level actors when using AI policies.

As we also demonstrate below, our framework allows us to analyze outcomes under many possible definitions of social welfare.
To explore the space of welfare configurations, we manipulate the ratio of $\frac{\healthindexweight}{1-\healthindexweight}$ relative to the default calibration.
For example, to explore the case where \USState{} $i$ cares twice as much about health outcomes, we set $\healthindexweight_i$ such that $\frac{\healthindexweight_i}{1-\healthindexweight_i} = 2 \cdot \frac{\hat{\healthindexweight}_i}{1-\hat{\healthindexweight}_i}$, where $\healthindexweight_i$ is the health priority parameter used to compute $\socialwelfare_i$ during training and analysis, and $\hat{\healthindexweight}_i$ is the value obtained during calibration.
To simplify notation, we use $m_i$ to denote this relative re-scaling, where, for example, $m_i=3$ denotes that $\frac{\healthindexweight_i}{1-\healthindexweight_i} = 3 \cdot \frac{\hat{\healthindexweight}_i}{1-\hat{\healthindexweight}_i}$, i.e. that \USState{} $i$ cares three times as much about health outcomes.

Figure~\ref{fig:improved-social-welfare} (middle) shows welfare improvements using $m_p=4$.
In this case, the \Federal{} government's increased prioritization on health leads to very large subsidy levels (\$4.4T on average), which considerably improve state-level welfare.
These large subsidies induce a shift towards higher stringency in the state-level policies (and, hence, better health outcomes); hence, the federal objective (when $m_p=4$) benefits from this extreme subsidy level as well.

When the increased health priority is applied to the states ($m_{1:N}=4$), we again see consistent social welfare improvements compared to real-world policies (Figure~\ref{fig:improved-social-welfare}, bottom).
In this case, the improvement is driven simply from the capacity of the AI policies to adapt to different objectives, while the real-world stringency policies are (as expected) suboptimal for this parameterization of social welfare.

\subsection{Outcomes under Varying Welfare Objectives}

\begin{figure}[t!]
\centering
    \includegraphics[width=\linewidth]{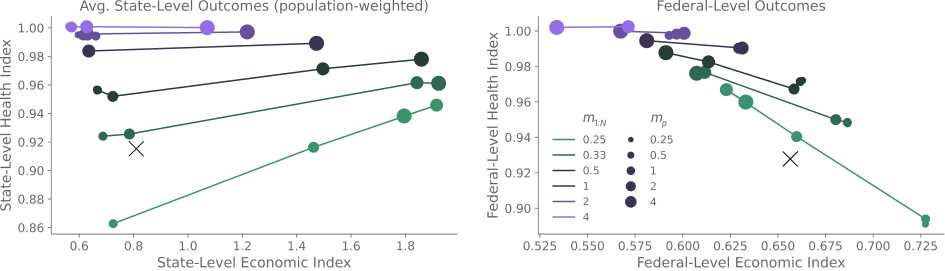}
    \caption{
        \textbf{Health and economic indices for different social welfare objectives.}
        Colors indicate the relative health priority scaling (defined in main text) $m_{1:N}$ of \USState{} policies.
        Dot sizes indicate the relative health priority scaling $m_p$ of \Federal{} policy.
        Black `X' denotes index values achieved under the real-world policy.
        \emph{Ceteris paribus}, as \USState{}s emphasize health more (higher $m_{i:N}$, fixed $m_p$), \USState{}-level health indices increase on average, but the economic index decreases.
        Similarly, as \Federal{} policy emphasizes health more (higher $m_p$), \USState{} and \Federal{} health indices increase, but the \Federal{} economic index decreases, reflecting the higher total borrowing cost of subsidies.
    }
    \label{fig:index-plots}
\end{figure}

The \textit{AI Economist} framework can flexibly optimize policy for any quantifiable objective.
Hence, we can explore the space of health and economic outcomes by changing the balance between health and economic policy objectives.

To demonstrate this flexibility, we train AI policies across a range of welfare parameterizations $\alpha_i$ at the \USState{}-level and \Federal{}-level.
We examine different re-scalings $m_i$ of the ratio $\frac{\hat{\healthindexweight}_i}{1-\hat{\healthindexweight}_i}$, where $m_i > 1$ captures that actor $i$ gives more weight to public health compared to the default calibration $\hat{\healthindexweight}_i$ (described above).
Figure~\ref{fig:index-plots} shows the state-level (left) and \Federal{}-level (right) outcomes, in terms of Health and Economic Indices, under various settings for $m_{1:N}$ (relative health prioritization of each \USState{}) and $m_p$ (relative health prioritization of the \Federal{} government).

As expected, changing the policy objective leads the AI policies to select a different trade-off between public health and the economy.
Higher $m_{1:N}$ lead to a higher Health Index at the expense of the Economic Index.
A similar trend is seen at the \Federal{} level with increasing $m_p$.
In our model, subsidies incentivize a \USState{} to be more stringent by reducing the economic burden of additional unemployment.
For certain settings of $m_{1:N}$ and $m_p$, the \Federal{} government prefers to use this economically costly mechanism to better align states' incentives with its own policy objective.
As a result, the trade-off between public health and the economy selected at the \Federal{} level depends on $m_p$.
Interestingly, however, because subsidies reduce the trade-off \USState{}s face between public health and the economy, higher $m_p$ tends to increase \textit{both} Health and Economic Indices at the state level.
However, this comes at a higher total borrowing cost to fund subsidies.

\subsection{Examining Learned Weights of the State-Level Stringency Policy}

\begin{figure}[t!]
\centering
    \includegraphics[width=\linewidth]{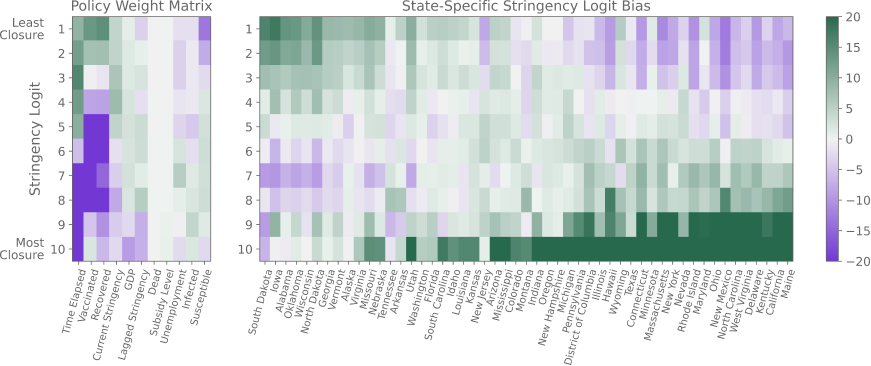}
    \caption{
        \textbf{Learned weights of the \USState{}-level policies.}
        Left: weights $W_{kj}$ for each input feature (columns indexed by $k$) and stringency level (rows indexed by $j$) show how learned \USState{} policies respond to pandemic conditions.
        For instance, as infections increase, stringency levels increase because $W_{kj} > 0$ for higher stringency levels and $W_{kj} < 0$ for lower stringency levels.
        Analogously, as recoveries increase, stringency levels decrease.
        Right: \USState{}-specific biases $b_{ij}$ show how \USState{}s (columns indexed by $i$) have varying stringency level preferences.
    }
    \label{fig:linear-weights}
\end{figure}

As discussed in Section~\ref{sec:frameowrk-features}, AI-driven policy design should emphasize explainability and simplicity when possible.
This motivated our choice of using log-linear policies.

Figure~\ref{fig:linear-weights} illustrates the learned stringency policy weights $W_{kj}$ and $b_{ij}$ (Eq.~\ref{eq:stringency-policy}), trained with health prioritizations $\hat{\alpha}_i$ and averaged over 10 repetitions with random seeds.
Examining $W$ confirms several intuitions: the learned policy is more likely to use higher stringency levels as susceptible and infected numbers increase.
Analogously, they are more likely to use lower stringency levels as recovered and vaccinated numbers increase.
In addition, the biases $b$ show how stringency level preferences vary across \USState{}.

Importantly, explainable policy model classes facilitate iterating policy design in practice by identifying potentially \emph{undesirable} input or policy features.
For example, we can see that $W$ encodes a strong shift towards lower stringency as the time index grows higher (``time elapsed'' in the Figure). However, timing indicators may be less semantically relevant than epidemiological or economic metrics.
Hence, a practitioner may decide to regularize AI policies to rely less or not at all on the time index or other input features.

\subsection{Sensitivity Analysis}

\begin{figure}[t!]
\centering
    \includegraphics[width=\linewidth]{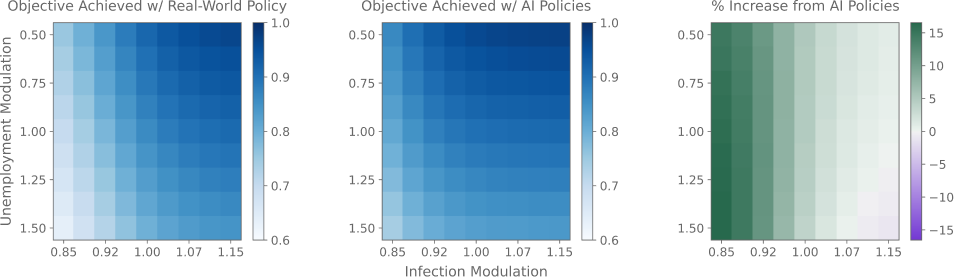}
    \caption{
        \textbf{Analyzing the impact of potential simulation errors.}
        Left: \Federal{}-level social welfare $\socialwelfare_p$ under real-world policies, across a range of model perturbations affecting how unemployment (rows) and COVID-19 transmission rates (columns) respond to stringency policy.
        Perturbations modulate the scale of the response; for example, with unemployment modulation $> 1$, the same increase in stringency yields a larger increase in unemployment (see Methods Section~\ref{sec:methods-sensitivity-analysis} for details).
        Middle: same, but under AI policies.
        Right: $\socialwelfare_p$ under AI policies, as a percentage of $\socialwelfare_p$ under real-world policies.
        Social welfare improvements persist across a large range of perturbations.
    }
    \label{fig:sensitivity}
\end{figure}
A natural concern with using simulation-trained AI policies to recommend real-world policy is that AI policies may \textit{overfit} to the simulated environment or fail to be robust to \emph{simulation-to-reality} gaps.
A \textit{sensitivity analysis} can test the robustness of AI policies under perturbations of the simulation parameters.
Such perturbations are a proxy for model fitting errors, noisy data, and other issues that cause model miscalibration and, potentially, model misspecification.

To illustrate this, we analyze how social welfare changes under perturbations of the unemployment and COVID transmission-rate model parameters.
Figure~\ref{fig:sensitivity} shows the \Federal{}-level social welfare $\socialwelfare_p$ over a grid of possible perturbations.
As expected, social welfare decreases for stronger unemployment rate and weaker transmission rate responses to changes in stringency level.
This is true both for real-world policies (Fig.~\ref{fig:sensitivity}, left) and AI policies (Fig.~\ref{fig:sensitivity}, middle), but, for the latter, social welfare is more robust to perturbations.
Gains from AI policies remain positive under most perturbation, except when unemployment and transmission-rate responses are strongly modulated (Fig.~\ref{fig:sensitivity}, right).
The stringency and subsidy choices of AI policies differ under each perturbation setting because these perturbations affect the inputs to the policy networks.
This analysis shows that the learned AI policy continues to yield well-performing policy recommendations across a space of simulation perturbations.

%% file: src/v2/discussion.tex
\section{Discussion}
\label{sec:Discussion}
Our case study demonstrated that the AI Economist framework has strong potential for real-world policy design.
We discuss some salient aspects and extensions of our framework.

\paragraph{Data Availability Constrains Simulation Model.}
All modeling choices are informed by the availability of data or existing domain knowledge (or lack thereof).
For example, we summarize pandemic response policy using a stringency level, as fine-grained details on the efficacy of individual policy levers is lacking.
Hence, AI policy design can be improved with more and higher-quality data which would enable more fine-grained design choices, e.g., enable including more health and economic variables.

\paragraph{Definition of Social Welfare.}
We defined social welfare as to capture the basic trade-off between public health and the economy.
In general, social welfare can include any number of priorities which are outside the scope of this work.
One could add, e.g., keeping ICU beds available, preventing businesses from failing, or minimizing inequality.
Applying the AI Economist in the real world would require robust consideration of how social welfare is defined and input from a sufficient set of representative stakeholders.

\paragraph{Retrospective Analysis uses Future Data to Emulate Domain Knowledge.}
We apply AI policies from the start of the COVID-19 pandemic, i.e., March 2020 onward, but we train AI policies in a simulator calibrated on data from all of 2020 \emph{that was gathered under the real-world policy}.
Hence, our analysis uses data that would not have been available in the real world, if hypothetically performed in March 2020.
Our use of ``future data'' emulates the use of 1) scientific estimates of unknowns, 2) domain knowledge, and 3) previous experience.
Together, such knowledge could provide forecasts of the pandemic and simulation parameter estimates, which in turn enable trainin AI policies.

\paragraph{Coarse Level of Data Aggregation.}
Policies may affect some social groups differently than others, but differential data on policy outcomes are not readily available for the COVID-19 pandemic.
Hence, it is hard to accurately simulate the effects of real-world diversity with the current data.
Our simulation models a version of the US where unemployment, the pandemic, and policy impact all people in the same way.
Representing diversity is a necessary step if this technology is eventually used in the real world.
However, this requires more robust fine-grained data than is readily  available.

\paragraph{Independency Assumptions across States.}
Our simulation uses an independent SIR model for each region and does not model interactions and cross-over effects between \USState{}s, i.e., that COVID-19 cases don't spread across \USState{} lines.
This simplifies calibration, while already demonstrating a good fit with real-world data.

\paragraph{Policy Impact Factors are Assumed Static.}
In our simulation, setting a particular stringency level will always lead to the same transmission rate or unemployment level.
In reality, the effect of policies may depend on past policy choices and the duration of the pandemic in complex ways.
For instance, it may be possible that after a long period of highly stringent policy and subsequent more relaxed policy, a second period of renewed stringent policy may not be as effective as public adherence may decrease due to fatigue.
The lack of fine-grained data makes it difficult to model such subtleties.
Adding such features could improve the realism of our simulation.

\paragraph{Structural Estimation, Correlation, Causation, and Generalization.}
In our model, the unemployment model and infection rates depend on the stringency level of \USState{} policy only.
This is an intuitive high-level modeling choice which yields strong out-of-sample performance given historical data.
However, there is no (counterfactual) data to ensure this is not a spurious correlation, nor that other causal factors may yield models that generalize better.
In particular, our model extrapolates the behavior of unemployment and transmission rates (as a function of policy measures) to pandemic situation that are unseen in the real world.
As such, real-world policy design entails continuous re-calibration and structural estimation.

\paragraph{Modeling Human Behavior using Machine Learning.}
We assume that all actors behave rationally, i.e., optimize policy for a given definition of social welfare.
However, in the real world actors may not behave optimally or their behavior may not be well explained by standard objective functions.
For example, many human cognitive biases are known, such as recency bias, ownership bias, behavioral inattention \autocite{gabaix2019behavioral}, and are well studied in behavioral economics \autocite{mullainathan2000behavioral}.

Hence, rational RL agents may not be sufficiently representative of the real world.
Moreover, multi-agent learning algorithms may not always represent how real-world actors learn and adapt in the presence of others,
or make unrealistic assumptions about how much actors know about the behavior of other actors.
As such, it would be fruitful to explore extensions of our framework that include human-like learning and behaviors.
However, it is an open challenge to collect sufficient micro-level data for these purposes.

\section{
    Ethics
}
This work should be regarded as a proof of concept. There are many aspects of the real world that our simulation does not capture. We do not endorse using the AI policies learned in this simulation for actual policymaking.
For an extended ethics review, see \url{https://blog.einstein.ai/ai-economist-data-driven-interpretable-robust-policy}.

%% file: src/v2/methods.tex
\section{Methods}

\subsection{Notation and Definitions}\label{sec:methods-notation-and-definitions}
We consider two types of actors:
\begin{itemize}
\item $N = 51$ \USState{} governors, indexed by $i \in 1,\ldots,N$, including Washington D.C., and
\item the \Federal{} government, indexed by $i = p$
\end{itemize}
Each \USState{} $i$ has a population of $\population_i$ people.
The US has a population of $\sum_{i} \population_i$ people.
Actors are indexed by $i = 1,\ldots, N, p$.
For a full overview of used variables, see Table \ref{ext-data-tab:model-parameters} and Table \ref{ext-data-tab:rl-parameters}.
\subsection{Simulation Dynamics}\label{sec:methods-simulation-dynamics}%
Our simulation aims to capture the impact of public health policy on the spread of COVID-19 and on the economy across the US.
Based on available data, we model (separately for each \USState{}) the spread of COVID-19 using a SIR model~\autocite{kermack_contribution_1927} and model economic output via unemployment and direct payments provided by the \Federal{} government.

\paragraph{Epidemiology Model.}
The SIR model is a standard epidemiological model that describes the dynamics of an infectious disease.
It subdivides a population $\population_i$ into 3 groups: susceptible $\susceptible_{i,t}$, infected $\infected_{i,t}$, and recovered $\recovered_{i,t}$.
The subscript $i \in \{1, \ldots, N \}$ denotes a \USState{}.
The subscript $t$ denotes time, where each timestep represents a single day.
Susceptible individuals become infected (transition from $\susceptible$ to $\infected$) at a rate that depends on the number of infected individuals and the (policy-dependent) transmission rate $\infectionrate_{i,t}$.
Infected individuals ``recover'' (transition from $\infected$ to $\recovered$) at a rate of $\recoveryrate$, which is the same across all \USState{}s.
By convention, $\recovered$ includes deaths $\deaths_{i,t}$: ``recovered'' individuals die at a rate of $\mortalityrate$, also the same across all \USState{}s.

To model vaccinations, we add an additional vaccinated group $\vaccinated_{i,t}$.
Susceptible individuals that receive a vaccine transition directly from $\susceptible$ to $\vaccinated$ and therefore cannot contract, spread, or die from the disease. This is a simplifying assumption, although it is straightforward to extend this model to the case where vaccinated people have a (much) lower chance of contracting and spreading the disease.

The SIR and vaccination dynamics are captured in the following equations at the \USState{}-level:

\eq{\label{eq:sir-equations-first}
  \susceptible_{i,t} &= \susceptible_{i,t-1} - \infectionrate_{i,t}\frac{\susceptible_{i,t-1}\infected_{i,t-1}}{\population_{i}}
  - \Delta\vaccinated_{i,t}
  \\
  \infected_{i,t} &= \infected_{i,t-1} + \infectionrate_{i,t} \frac{\susceptible_{i,t-1}\infected_{i,t-1}}{\population_{i}} - \recoveryrate \infected_{i,t-1} \\
  \recovered_{i,t} &= \recovered_{i,t-1} + \recoveryrate \infected_{i,t-1} \\
  \vaccinated_{i,t} &= \vaccinated_{i,t-1} + \Delta\vaccinated_{i,t} \\
  \deaths_{i,t} &= \mortalityrate \cdot \recovered_{i,t} \\
  \label{eq:sir-equations-last}
  \population_{i} &= \susceptible_{i,t} + \infected_{i,t} + \recovered_{i,t} + \vaccinated_{i,t}.
}
Note that the transmission rate $\infectionrate_{i,t}$ is location- and time-specific; this reflects differences between \USState{}s and the stringency level of each State's public health policy (which can vary over time).
Furthermore, following the standard SIR model, we do not include population growth: the population $\population_i$ is fixed and does not depend on time.
Moreover, we do not model infections between \USState{}s, e.g., infection through neighbors or airplane routes.

We model the transmission rate of \USState{} $i$ at time $t$ as a linear function of the State's stringency level $\stringencyindex_{i,t-d}$, delayed by $d$ days:
\eq{\label{eq:stringency-to-transmission}
    \infectionrate_{i,t} = \infectionrateslope_i \cdot \stringencyindex_{i,t-d} + \infectionratebias_i,
}
where $\infectionrateslope_i$ and $\infectionratebias_i$ are the slope and intercept of the \USState{}-specific linear function.

$\Delta\vaccinated_{i,t}$ denotes new daily vaccinations.
The onset of vaccines is delayed until the simulation has reached a specified date $\vaccinationstartdate$.
After that date, a fixed amount of vaccines are dispersed daily to each \USState{}, that amount being proportional to the population size $\population_i$, at a rate $\vaccinationrate_i$.
Specifically:
\eq{
  \Delta\vaccinated_{i,t} = \dailyvaccinationrate_i \cdot \bm{1}[t \geq \vaccinationstartdate], \quad \dailyvaccinationrate_i = \vaccinationrate_i \population_i, \quad
  0 < \vaccinationrate_i < 1,
}
where $\bm{1}$ is the indicator function.

\paragraph{Economic Model.}
Economic output is modeled via unemployment and direct payments (i.e. subsidies) provided by the \Federal{} government.
Unemployment in \USState{} $i$ at time $t$ is modeled by convolving the history of stringency level changes with a bank of $K$ exponentially-decaying filters:
\eq{
    \tilde{\unemployment}_{i,t}^k &= \sum_{t'=t-L}^{t'=t} e^{\frac{t'-t}{\lambda_k}} \cdot \Delta \stringencyindex_{i, t'}, \\
    \unemployment_{i,t} &= \texttt{softplus}\brck{\sum_{k=1}^K w_{i, k} \cdot \tilde{\unemployment}_{i,t}^k} + \unemployment^0_i.
}
The term $\tilde{\unemployment}_{i,t}^k$ denotes the unemployment response captured by filter $k$, which has decay constant $\lambda_k$,
$L$ is the filter length, and $\Delta \stringencyindex_{i, t}$ is the change in stringency level in \USState{} $i$ at time $t$.
Each \USState{}'s excess unemployment at time $t$ is computed as a linear combination of $\tilde{\unemployment}_{i,t}$ using \USState{}-specific weights $w_i$ and a \texttt{softplus} function to ensure that excess unemployment is positive.
Finally, unemployment $\unemployment_{i, t}$ is taken as the sum of excess unemployment and baseline unemployment $\unemployment_i^0$.

We calibrate the daily economic output per employed individual such that, at baseline levels of unemployment, total yearly GDP of the simulated US matches the actual US GDP in 2019.
Each \USState{}'s daily economic output $\productivity_{i,t}$ is modeled as the total output of its working population $\omega_{i,t}$ at time $t$ plus any money provided via direct payments from the \Federal{} government at that time $\statesubsidy_{i, t}$.
Concretely, the number of available workers is
\eq{
\omega_{i,t} = \nu \left( \population_i - \deaths_{i,t} - \eta\cdot\infected_{i,t} \right ) - \unemployment_{i,t}.
}
And daily economic output is therefore
\eq{
\productivity_{i,t} = \kappa \cdot \omega_{i,t} + \statesubsidy_{i, t}.
}
Here, $\nu=0.6$ captures the portion of the population that is working age and $\kappa$ captures the average daily economic output per active worker.
Note that individuals that have died, as well as a fraction ($\eta=0.1$) of infected individuals, are accounted for in determining the number of available workers.

\subsection{Data and Model Calibration}
\label{sec:methods-data-and-model-calibration}
Our simulation is grounded in real-world data.
We combine publicly available data on State-by-State COVID-19 deaths, unemployment, and stringency of public health policy, as well as \Federal{} policy concerning direct payments.

\paragraph{Policy Data.}
We take the daily stringency estimates provided by the Oxford COVID-19 Government Policy Tracker~\autocite{Hallas2021}, discretized into 10 levels, as the real-world data for $\stringencyindex$.
The date and amount of each set of direct payments issued through \Federal{} policy are taken from information provided by the COVID Money Tracker project~\autocite{CovidMoneyTracker}.

\paragraph{Public Health Data.}
To fit the disease model, we use the daily cumulative COVID-19 death data provided by the COVID-19 Data Repository at Johns Hopkins University~\autocite{Dong2020}.
We treat death data $\deaths_{i,t}$ as ground-truth and solve for $\susceptible_{i,t}, \infected_{i,t}, \recovered_{i,t},$ and $\infectionrate_{i,t}$ algebraically using the SIR equations (Eqs.~\ref{eq:sir-equations-first}-\ref{eq:sir-equations-last}), given fixed estimates of the mortality rate $\mortalityrate=0.02$ and recovery rate $\recoveryrate=\frac{1}{14}$.
Estimating the missing data this way simply requires rearranging the SIR equations to express the unknown quantities in terms of ``known'' quantities such as $\deaths_{i,t}$, $\vaccinated_{i,t}$, $\mortalityrate$, and $\recoveryrate$.
For example, $\recovered_{i,t} = \frac{\deaths_{i,t}}{\mortalityrate}$
and $\infected_{i,t}=\frac{\recovered_{i,t+1}-\recovered_{i,t}}{\recoveryrate}$, and so on.

The SIR estimates are useful for setting the starting conditions of the simulation; however, this inference procedure primarily serves to estimate each state's daily transmission rate $\infectionrate_{i,t}$.
This allows us to measure the relationship between (delayed) stringency $\stringencyindex_{i,t-d}$ and COVID-19 transmission.
We set $d$ to 29 days, since the empirical correlation between stringency and transmission is strongest at this delay.

Finally, we use linear regression to fit the infection rate parameters of Equation~\ref{eq:stringency-to-transmission} for each state, while regularizing the state-to-state variability in these parameters to help prevent individual states from overfitting to noise.

We set the onset date for vaccinations and their daily rate of delivery to approximately match aggregate vaccination trends in the US.
Specifically, vaccines become available in the simulation after January 12th, 2021, at a rate of 3k new daily vaccines per 1M people.

\paragraph{Economic Data.}
Monthly unemployment rates for each \USState{} are collected from the Bureau of Labor Statistics~\autocite{BureauofLaborStatistics2021}; in the daily representation of these data, each day in a given month uses the reported value for that month.
We fit the unemployment parameters $\lambda_k$, $w_{i,k}$, and $\unemployment_i^0$ (for each $i$, $k$) by minimizing the squared error between predicted daily unemployment rates and those contained in the data.
As with the fit to transmission rate, we regularize the variability in $w_{i,k}$ across \USState{}s.

\paragraph{Train-Test Splits.}
The simulation is calibrated on data from 2020.
We use data through November 2020 to estimate model parameters and the remaining 2020 data to tune hyperparameters.
In particular, we confirmed that, when applying the real-world stringency policies, the calibrated simulation predicts deaths and unemployment level that are consistent with real-world outcomes throughout 2020 and through April 2021.

Within the 2020 calibration data, predicted outcomes capture 80\% and 39\% of the population-normalized variance in unemployment and COVID-19 deaths, respectively.
For the 2021 ``test'' data, the predictions capture 54\% of unemployment variance and 35\% death variance.

\subsection{Social Metrics and Indices}\label{sec:methods-social-metrics-and-indices}
Grounding the simulation in real-world data reveals a tradeoff between minimizing the spread of COVID-19 and minimizing economic hardship, such as unemployment.
To quantify health and economic outcomes under various policy choices, we define a Health Index $\healthindex$ and Economic Index $\economicindex$.
Our framework does not place any constraints on how social welfare $\socialwelfare_i$ is defined, but, for the purposes of this study, we define it to be a weighted sum of the Health Index and the Economic Index:
\eq{\label{supp:eq:socialwelfare}
   \socialwelfare_i = \socialwelfaremixingcoeff_i\healthindex_i + \left(1-\socialwelfaremixingcoeff_i\right)\economicindex_i,
}
where $0 \leq \socialwelfaremixingcoeff_i \leq 1$ is a mixing parameter that captures the relative prioritization of health outcomes over economic outcomes for actor $i$.

The episode-averaged index values are the average of \emph{normalized marginal index values} throughout the simulation episode:
\eq{
    \healthindex_{i} \triangleq \frac{1}{T} \sum_{t=1}^T \Delta\healthindex_{i, t},
    \qquad
    \economicindex_{i} \triangleq \frac{1}{T} \sum_{t=1}^T \Delta\economicindex_{i, t}.
}

Intuitively, the Health Index decreases with the number of deaths and the Economic Index increases with overall economic output.
We denote the unnormalized, marginal Health and Economic Index of \USState{} $i$ at time $t$ as $\Delta\tilde{\healthindex}_{i, t}$ and $\Delta\tilde{\economicindex}_{i, t}$, respectively.
They are defined as:
\eq{
    \Delta\tilde{\healthindex}_{i, t} = -\Delta\deaths_{i, t},
    \qquad
    \Delta\tilde{\economicindex}_{i, t} = \crra \left( \frac{\productivity_{i, t}}{\productivity_i^0} \right).
}
The baseline productivity $\productivity_i^0$ is the daily economic output expected under baseline, i.e., non-pandemic, conditions.
We include a CRRA nonlinearity~\autocite{arrow1971theory} in the Economic Index, where $\crra(x) = 1 + \frac{x^{1-\eta}-1}{1-\eta}$ and $\eta=2$ is a shape parameter.
As a result, there are diminishing marginal returns on economic output.\footnote{
The motivations for this modeling choice are discussed in the main text.
}

Indices for the \Federal{} government are defined similarly, but sum over the entire country and reflect the borrowing cost of any direct payments:
\eq{
    \Delta\tilde{\healthindex}_{p, t} = -\sum_{i=1}^N\Delta\deaths_{i, t},
    \qquad
    \Delta\tilde{\economicindex}_{p,t} = \crra \left( \frac{\sum_{i=1}^N \productivity_{i, t} - \costofborrowing \cdot \statesubsidy_{i, t}}{\sum_{i=1}^N \productivity_i^0} \right),
}
where $\costofborrowing \geq 1$ is the borrowing cost of providing \$1 of direct payments.
Note that $\productivity_{i,t}$ already includes money received through direct payments $\statesubsidy_{i, t}$, so additional payments will always increase the Economic Index at the \USState{} level and decrease it at the \Federal{} level.

To standardize analysis, we normalize the marginal indices based on their minimum and maximum values, giving normalized marginal indices $\Delta\healthindex_{i, t}$ and $\Delta\economicindex_{i, t}$ for each actor $i$, as:
\eq{
    \Delta\healthindex_{i, t} = \frac{\Delta\tilde{\healthindex}_{i, t} - \Delta\healthindex_i^\textrm{min}}{\Delta\healthindex_i^\textrm{max} - \Delta\healthindex_i^\textrm{min}},
    \qquad
    \Delta\economicindex_{i, t} = \frac{\Delta\tilde{\economicindex}_{i, t} - \Delta\economicindex_i^\textrm{min}}{\Delta\economicindex_i^\textrm{max} - \Delta\economicindex_i^\textrm{min}}.
}
We obtain these minimum and maximum values from the average marginal indices measured by running the simulation under 2 policy extremes: minimum-stringency (i.e., fully-open) and maximum-stringency (i.e., fully-closed).
The minimum-stringency policy contributes the minimum and maximum marginal Health and Economic indices, respectively, and vice versa for the maximum-stringency policy.
When normalizing this way, each minimum-stringency policy yields $\healthindex_i= 0$ and $\economicindex_i= 1$ and each maximum-stringency policy yields $\healthindex_i= 1$ and $\economicindex_i= 0$.
As a result, higher index values denote more preferable outcomes.

Each \USState{} and \Federal{} actor aims to maximize the social welfare $\socialwelfare_i$ of its jurisdiction.
Therefore, when applying RL, the instantaneous reward function is therefore the weighted sum of the normalized
marginal indices:
\eq{\label{eq:instantaneous_reward_agent}
   \rew_{i,t} = \socialwelfaremixingcoeff_i\Delta\healthindex_{i,t} + \left(1-\socialwelfaremixingcoeff_i\right)\Delta\economicindex_{i,t}.
}

\subsection{Calibrating Policy Priorities from Data}\label{sec:methods-calibrating-policy-priorities-from-data}
To facilitate comparison between real-world policies and policies learned in our framework, we optimize AI policies for the social welfare with health priority parameters $\socialwelfaremixingcoeff_i$ that best fit real-world policies.
Of course, it is very hard to know the exact mathematical form of the objectives of real-world actors.
Rather, we attempt to identify the parameters $\hat{\socialwelfaremixingcoeff}, \hat{\socialwelfaremixingcoeff}_i$ that best explain the outcomes achieved by the real-world policy, assuming they used the social welfare objective as defined in Equation \ref{supp:eq:socialwelfare}.

To obtain this estimate, we first collect simulated health/economic outcomes under 3 policies: the actual stringency choices, minimum stringency, and maximum stringency.
We use these outcomes to estimate the Pareto frontier in the $\left( \healthindex_i, \economicindex_i \right)$ coordinate space.
By definition, the $\left( \healthindex_i, \economicindex_i \right)$ coordinates for the minimum- and maximum-stringency policies define the endpoints of this frontier, at $\left(0, 1\right)$ and $\left(1, 0\right)$, respectively.
We assume the Pareto frontier for state $i$ has form $\economicindex_i = (1-\healthindex_i)^{x_i}$ and that the coordinates associated with the actual-stringency policy are found along this frontier.
We set the shape parameter $x_i$ based on this latter assumption, and take $\hat{\socialwelfaremixingcoeff}_i$ as the value that maximizes social welfare along the estimated Pareto frontier:
\eq{
  \hat{\socialwelfaremixingcoeff} =
  \max_{\socialwelfaremixingcoeff_i} \socialwelfaremixingcoeff_i\healthindex_i(\bm{\policy}) + \left(1-\socialwelfaremixingcoeff_i\right)\economicindex_i(\bm{\policy}, \healthindex_i) =
  \max_{\socialwelfaremixingcoeff_i} \socialwelfaremixingcoeff_i\healthindex_i(\bm{\policy}) + \left(1-\socialwelfaremixingcoeff_i\right)(1-\healthindex_i(\bm{\policy}))^{x_i},
}
where the Economic and Health Index values are obtained from running the policies $\bm{\policy}$ in the simulation.
In other words, given the estimate of the Pareto frontier, we find the $\hat{\socialwelfaremixingcoeff}_i$ that best rationalizes the outcomes obtained under the actual policy, i.e. the $\hat{\socialwelfaremixingcoeff}_i$ under which these outcomes are considered optimal.

When obtaining these estimates, we simulate the same period of time used for calibrating the simulation.
As such, we calibrate $\hat{\socialwelfaremixingcoeff}_i$ from outcomes measured after running the simulation from March 23, 2020 to December 31, 2020.

Assuming a functional form for the social welfare objective and inferring $\hat{\socialwelfaremixingcoeff}$ for real-world policies facilitates comparing AI and real-world policies, and is not a complete view of what social welfare consistitutes in the real world.
We emphasize that future use of AI policy design frameworks would involve first specifying policy priorities, e.g., what social welfare means, as opposed to inferring them \emph{post-hoc}.

\subsection{Multi-Agent Reinforcement Learning}
\label{sect:rl}
We learn the optimal policy for all actors using multi-agent RL \autocite{sutton2018reinforcement}. We use the term \emph{actor} to refer to the two levels of agents: \USState{} governments (agents) and the \Federal{} government (social planner).
Actors learn by interacting with a simulation environment, which iterates between states using dynamics $\trans\brck{\St_{t+1} | \St_t, \Ac_t}$, where $\St_t$ and $\Ac_t$ denote the collective states and actions of the actors at time step $t$.
For each time step $t$, the social planner receives an observation $\ob_{p,t}$, executes an action $\ac_{p,t}$ and receives a reward $\rew_{p,t}$.
Similarly, each agent $i = 1,\ldots,N$ receives an observation $\ob_{i,t}$, executes an action $\ac_{i,t}$ and receives a reward $\rew_{i,t}$.
Note that each agent does not instantly observe action $\ac_{p,t}$ of the planner in its observation $\ob_{i,t}$. However, agents $i$ may see the effect of the planner's action $\ac_{p,t}$ at later times $t' > t$, e.g., if the planner increased subsidies for the next 90 days.
Once all actors have acted, the environment transitions to the next state $\st_{t+1}$, according to the transition distribution $\trans$.

\paragraph{Policy Models.}
Each actor learns a policy $\pol$ that maximizes its $\df$-discounted expected return.
We denote \USState{} policies as $\pol_i\brck{\ac_{i,t} | \ob_{i,t}; \theta_i}$ and \Federal{} policy as $\pol_p\brck{\ac_{p,t} | \ob_{p,t}; \theta_p}$.
Here, $\theta_i$ and $\theta_p$ parameterize the policies.
Let $\bm{\pol} = \brck{\pol_1,\ldots,\pol_N, \pol_p}$ denote the collection of all policies and $\bm{\pol}_{-j} = \brck{\pol_1,\ldots,\pol_{j-1},\pol_{j+1},\ldots,\pol_N, \pol_p}$ denote the joint policy without actor $j$.

Through RL, actor $j$ seeks a policy to maximize its expected reward:
\eq{\label{eq:rl-agent-objective}
\max_{\theta_j} \E_{a_j\sim \pi_j, \va_{-j}\sim\bm{\pi}_{-j}, \st'\sim\trans} \brcksq{ \sum_{t=0}^\eplen \df^t \rew_{j,t}},
}
with discount factor $\gamma \in (0, 1)$.
The expected reward depends on the behavioral policies $\bm{\pi}_{-j}$ of the other actors and the environment transition dynamics $\trans$.
As such, Equation \ref{eq:rl-agent-objective} yields an agent policy that best respond to the policies of other actors, given the dynamics of the simulated environment and the actor's observations.

\paragraph{Sharing Weights.}
For data efficiency, all $N$ \USState{} agents share the same parameters during training, denoted $\mweight$, but condition their policy $\policy_i\brck{\ac_i|\ob_i; \mweight}$ on agent-specific observations $\ob_i$, which includes their identity.
In effect, if one agent learns a useful new behavior for some part of the state space then this becomes available to another agent.
At the same time, agent behaviors remain heterogeneous because they have different observations.

\paragraph{Agent States and Actions.}
Each agent $i$ (representing the governor of \USState{} $i$) chooses a stringency level $\stringencyindex_i \in [1,\ldots,10]$, where $\stringencyindex_i=1$ and $\stringencyindex_i=10$ denote the minimum and maximum stringency levels, respectively.
The agent's observation is:
\eq{
  \ob_{i,t} = (
    i,
    \susceptible_{i,t},
    \infected_{i,t},
    \recovered_{i,t},
    \deaths_{i,t},
    \vaccinated_{i,t},
    \unemployment_{i,t},
    \productivity_{i,t},
    \stringencyindex_{i,t},
    \stringencyindex_{i,t-d},
    \statesubsidy_{i,t}
  ),
}
which includes a one-hot encoding of the agent's index, the states of the SIR components, unemployment, (post-subsidy) productivity, current and delayed stringency, and the current level of subsidy provided by the \Federal{} government (see below).

\paragraph{Planner States and Actions.}
The planner chooses a subsidy level $\pi_{p} \in [1,\ldots,20]$ that controls the amount of \textit{direct payments} provided to workers.
Each timestep, some amount of money is added directly to each \USState{}'s daily productivity $\productivity_{i,t}$, and that amount depends on the subsidy level and the State's population.
At the minimum subsidy level, no money is added.
At the maximum, each \USState{} receives a daily subsidy of roughly \$55 per person, corresponding to a direct payment rate of \$20k per person per year.
The planner observes:
\eq{
  \ob_{p,t} = (
    \susceptible_{t},
    \infected_{t},
    \recovered_{t},
    \deaths_{t},
    \vaccinated_{t},
    \unemployment_{t},
    \productivity_{t},
    \stringencyindex_{t},
    \stringencyindex_{t-d},
    \statesubsidy_{t}
  ).
}
The planner observes the same types of information as the agents, but for all \USState{}s at once.
For instance, $\susceptible_t$ denotes that the planner observes the 51-dimensional array of susceptible rates for each \USState{}.

\subsection{Constraining Policies for Improved Realism}

To improve realism and the real-world viability of learned policies, we constrain RL policies to change slowly, as follows:
\begin{itemize}
  \item We restrict the frequency with which each AI governor can update its stringency $\stringencyindex_i$.
  Specifically, if agent $i$ acts so as to change the stringency on timestep $t$, it is prevented from making any further changes until timestep $t+28$.
  In other words, the stringency level for a \USState{} can change at most once every 28 days.
  \item The planner updates the level of direct payments $\pi_p$ every 90 days.
\end{itemize}
Future research could explore other constraints on policies, e.g., only allowing one increase in stringency level, and subsequently only decreases.

\subsection{Two-level RL Strategies}

Following~\autocite{zheng2020ai,zheng2021ai}, we use \textit{entropy regularization}~\autocite{williams1991function} to stabilize simultaneous agent and planner optimization.
Specifically, we schedule the amount of entropy regularization applied to the planner policy such that the planner policy is essentially random during the early portion of training.
As a result, agents learn to conditioned policies on a wide range of possible subsidies while the planner's entropy regularization coefficient is gradually annealed to its final value.
After this annealing, both the agents and the planner are able to stably optimize their policies towards their respective objectives.

\hypertarget{Sensitivity Analysis}{%
\subsection{Sensitivity Analysis}\label{sec:methods-sensitivity-analysis}}

We perform a sensitivity analysis to examine how systematic errors in model calibration affect simulated outcomes.
In particular, this is useful to examine the possible range of calibration errors within which AI policies outperform real-world policies.
We examine sensitivity to systematic under-/over-estimations of two factors: how much (1) the COVID-19 transmission rate and (2) the excess unemployment respond to the stringency policy.
To simulate these conditions we replace $\infectionrate^{\stringencyindex}_i$ with $m^{\infectionrate} \cdot \infectionrate^{\stringencyindex}_i$ and replace $w_{i,k}$ with $m^w \cdot w_{i,k}$, where $m^{\infectionrate}$ and $m^w$ denote the amount of \textit{modulation} of the transmission rate response and unemployment response, respectively.
For example, $m^w=1.5$ emulates the condition that the \textit{actual} unemployment response is 50\% larger than in the default calibration.\footnote{For instance, if actual unemployment numbers are under-reported in the calibration data.}

We set $m^{\infectionrate}=1$ and $m^w=1$ during training.
When performing the sensitivity analysis, we collect simulated outcomes across a grid of $0.85 \leq m^{\infectionrate} \leq 1.15$ and $0.5 \leq m^w \leq 1.5$ values, for both the real-world policies and with the AI policies.
Note that the real-world policies are fixed, but the AI policies may change for different $m^{\infectionrate}$ and $m^w$ since these impact the observations on which the AI policy decisions are based.

%% file: src/v2/end-notes.tex
\hypertarget{endnotes}{%
\section{End Notes}\label{sec:End Notes}}
\begin{itemize}
\item \textbf{Acknowledgements.}
    We thank Yoav Schlesinger and Kathy Baxter for the ethical review.
    We thank Silvio Savarese for his support.
    We thank Lav Varshney, Andre Esteva, Caiming Xiong, Michael Correll, Dan Crorey, Sarah Battersby, Ana Crisan, Maureen Stone for valuable discussions.
\item \textbf{Author Contributions.}
SS, AT, and SZ contributed equally.
AT, SS, SH, and SZ developed the theoretical framework.
DW, SS, and AT cataloged and processed data.
SS, AT, and SZ developed the simulator.
SS and AT performed experiments.
AT, SS, SH, and SZ analyzed experiments.
AT and SZ drafted the manuscript.
SS and SH commented on the manuscript.
AT and SZ conceived the project.
SZ planned and directed the project.
\item Source code for the economic simulation is available at \url{https://www.github.com/salesforce/ai-economist}.
\item More information available at \url{https://www.einstein.ai/the-ai-economist}.
\item The authors declare no competing interests.
\item All data needed to evaluate the conclusions in the paper are present in the paper and/or the Supplementary Materials.
\item All data can be provided pending scientific review and a completed material transfer agreement. Requests should be submitted to \url{stephan.zheng@salesforce.com}.
\item The authors acknowledge that they received no funding in support for this research.
\end{itemize}

%% file: src/v2/extended-data.tex
\section{Extended Data}

\begin{figure}[H]
    \centering
        \includegraphics[width=\linewidth]{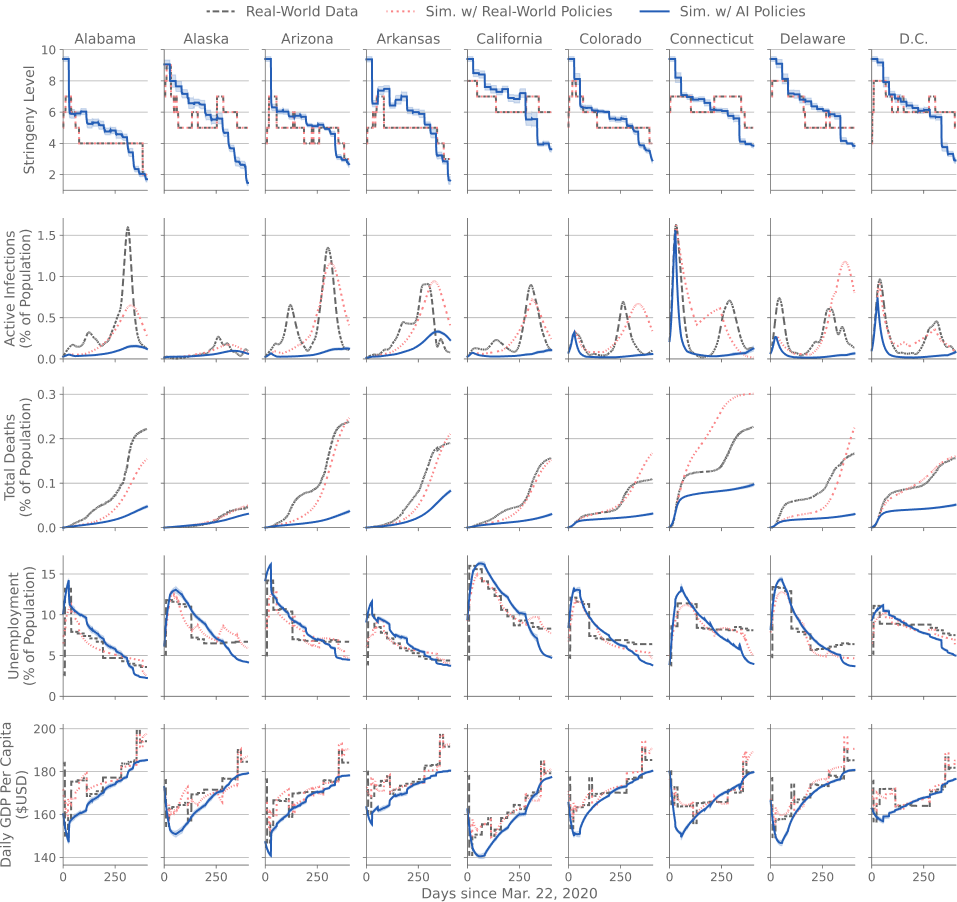}
        \caption{
            \textbf{Real-world vs simulation outcome at the state level (Alabama - D.C.).}
            Conventions are the same as Figure~\ref{fig:state-comparison} in Section~\ref{sec:Results} of the main text.
        }
        \label{fig:state-comparison-alabama-dc}
    \end{figure}

    \begin{figure}[H]
    \centering
        \includegraphics[width=\linewidth]{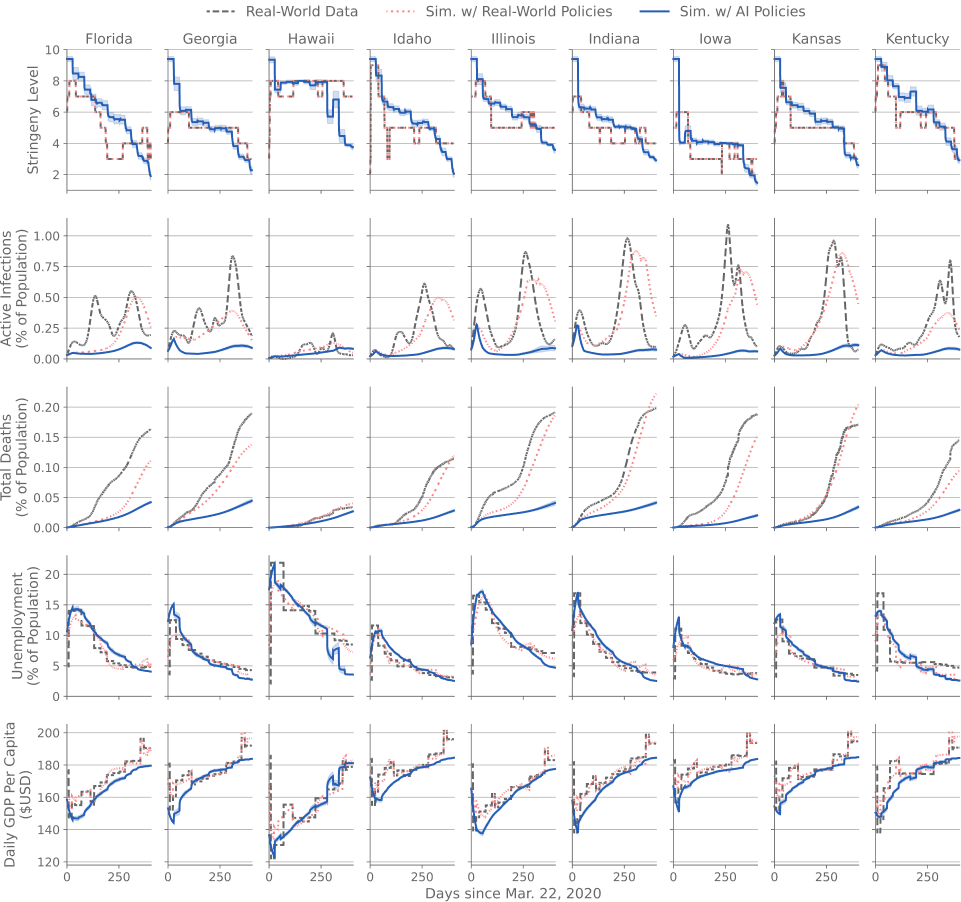}
        \caption{
            \textbf{Real-world vs simulation outcome at the state level (Florida - Kentucky).}
            Conventions are the same as Figure~\ref{fig:state-comparison} in Section~\ref{sec:Results} of the main text.
        }
        \label{fig:state-comparison-florida-kentucky}
    \end{figure}

    \begin{figure}[H]
    \centering
        \includegraphics[width=\linewidth]{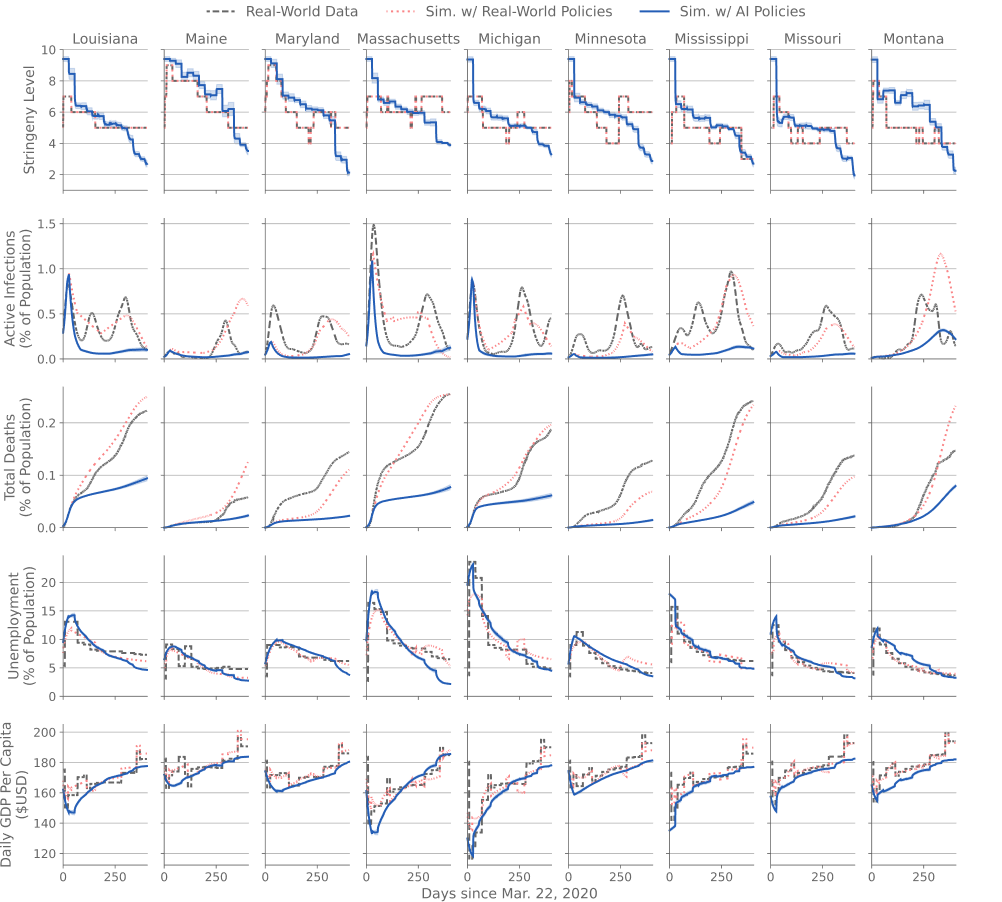}
        \caption{
            \textbf{Real-world vs simulation outcome at the state level (Louisiana - Montana).}
            Conventions are the same as Figure~\ref{fig:state-comparison} in Section~\ref{sec:Results} of the main text.
        }
        \label{fig:state-comparison-louisiana-montana}
    \end{figure}

    \begin{figure}[H]
    \centering
        \includegraphics[width=\linewidth]{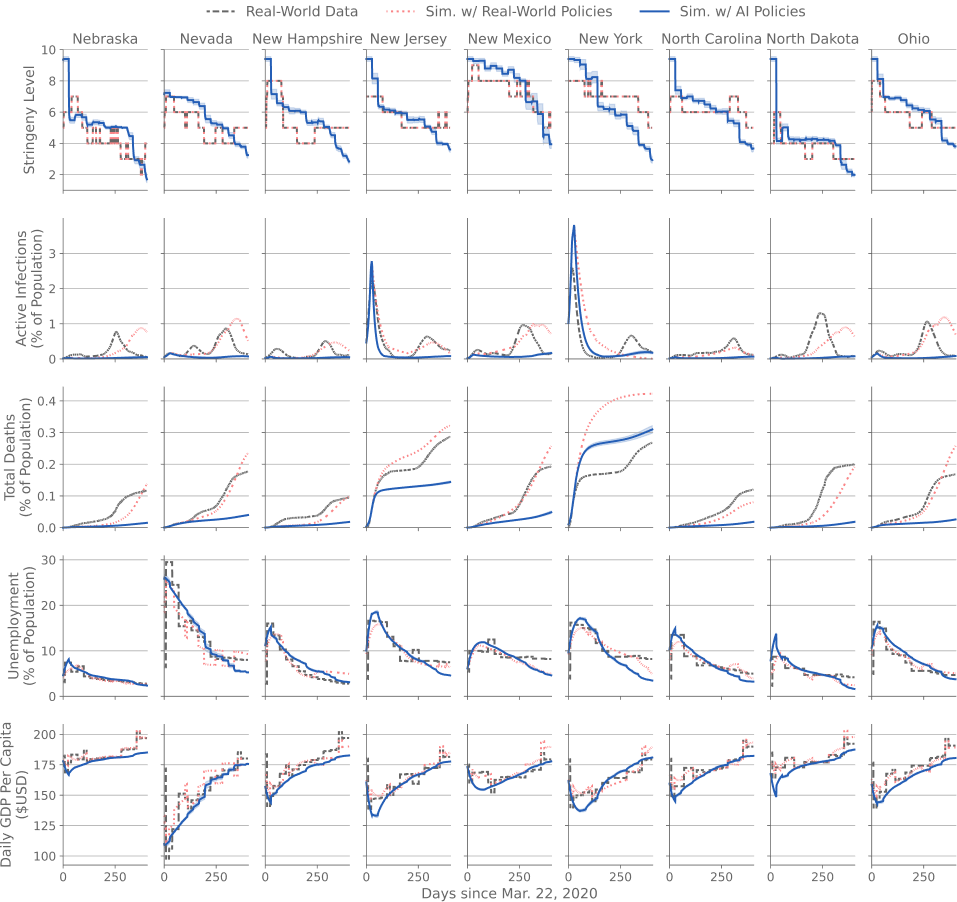}
        \caption{
            \textbf{Real-world vs simulation outcome at the state level (Nebraska - Ohio).}
            Conventions are the same as Figure~\ref{fig:state-comparison} in Section~\ref{sec:Results} of the main text.
        }
        \label{fig:state-comparison-nebraska-ohio}
    \end{figure}

    \begin{figure}[H]
    \centering
        \includegraphics[width=\linewidth]{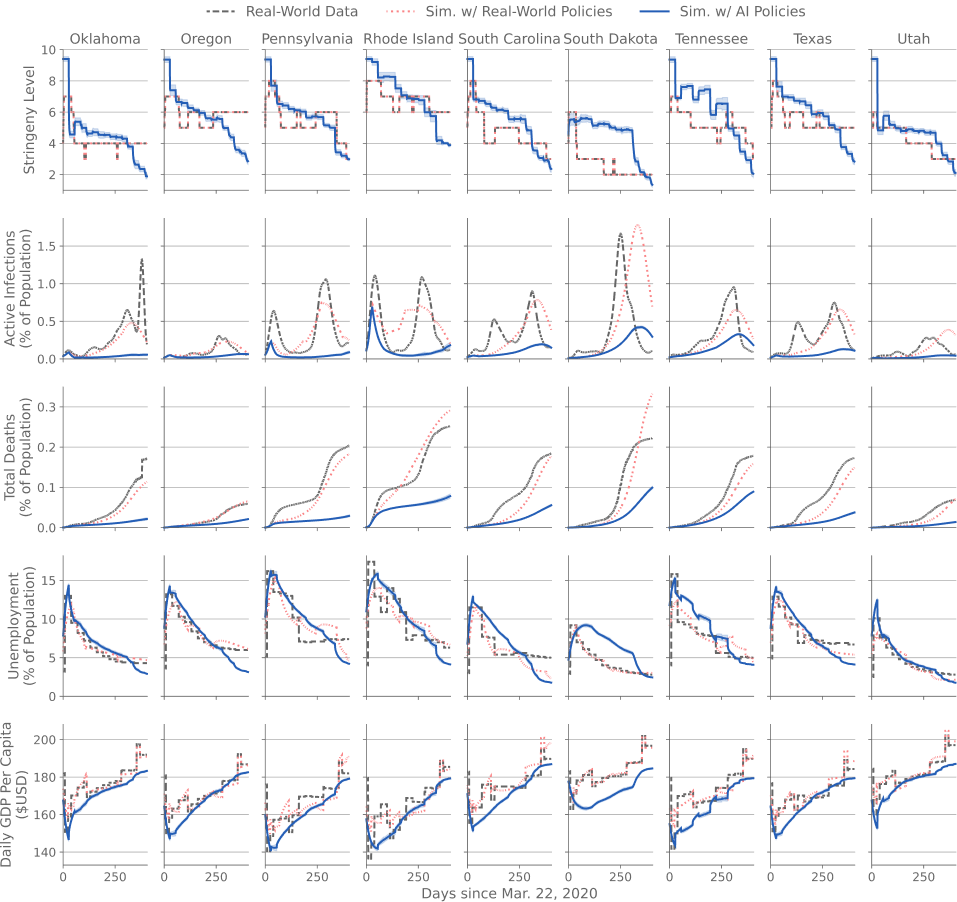}
        \caption{
            \textbf{Real-world vs simulation outcome at the state level (Oklahoma - Utah).}
            Conventions are the same as Figure~\ref{fig:state-comparison} in Section~\ref{sec:Results} of the main text.
        }
        \label{fig:state-comparison-oklahoma-utah}
    \end{figure}

    \begin{figure}[H]
    \centering
        \includegraphics[width=\linewidth]{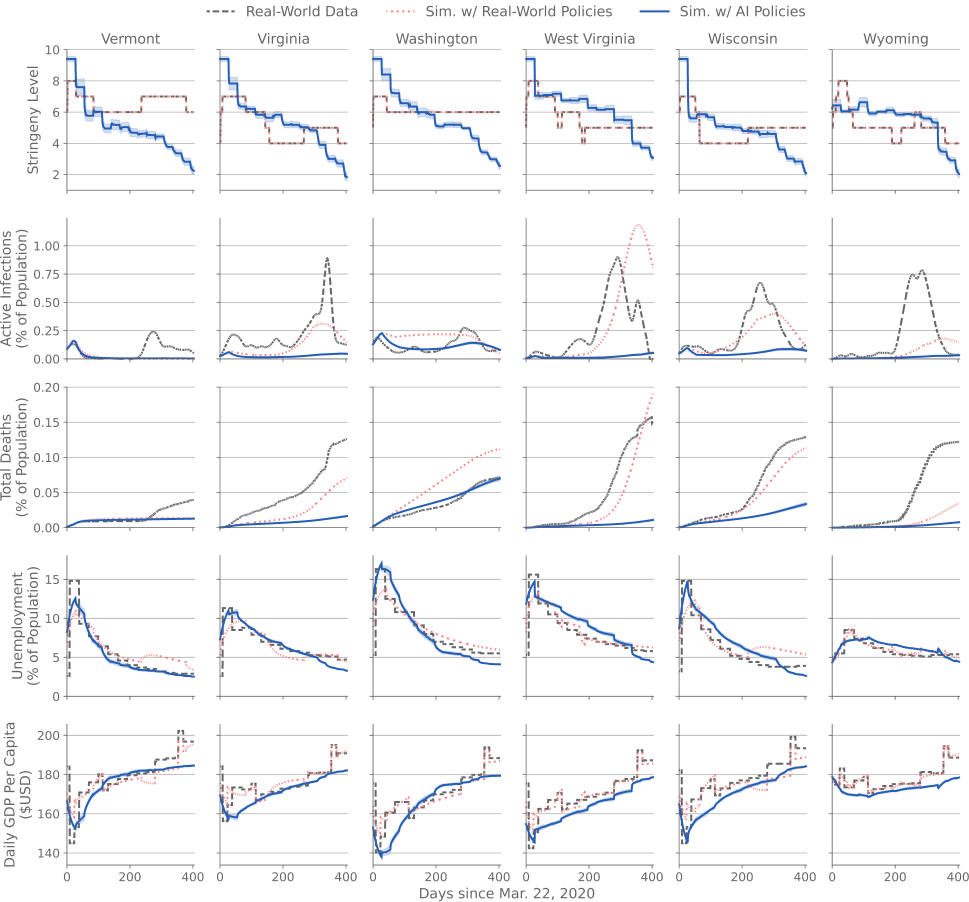}
        \caption{
            \textbf{Real-world vs simulation outcome at the state level (Vermont - Wyoming).}
            Conventions are the same as Figure~\ref{fig:state-comparison} in Section~\ref{sec:Results} of the main text.
        }
        \label{fig:state-comparison-vermont-wyoming}
    \end{figure}

\begin{table}[H]
    \begin{small}
        \begin{center}
            {\sffamily %
            \begin{tabular}[c]{lll}
                \hline
                Parameter & Symbol & Value \\
                \hline
                Population total for state $i$   &  $\population_i$  &  \\
                The number of people in state $i$ that are \textbf{susceptible} at time $t$   &  $\susceptible_{i,t}$   &  \\
                The number of people in state $i$ that are \textbf{infected} at time $t$      &  $\infected_{i,t}$  &  \\
                The number of people in state $i$ that are \textbf{recovered} at time $t$     &  $\recovered_{i,t}$ &  \\
                The number of people in state $i$ that are \textbf{vaccinated} at time $t$     &  $\vaccinated_{i,t}$ &  \\
                The number of people in state $i$ that have \textbf{passed away} at time $t$       &  $\deaths_{i,t}$  &  \\
                Infection rate   &  $\infectionrate_{i,t}$  &  \\
                Recovery rate   &  $\recoveryrate$  &  $\frac{1}{14}$ \\
                Mortality rate (the fraction of `recovered' people that pass away)   &  $\mortalityrate$  &  $2\%$ \\
                \hline
                Stringency index   &  $\stringencyindex_{i,t}$  &  \\
                Productivity   &  $\productivity_{i,t}$  &  \\
                State subsidy   &  $\statesubsidy_{i,t}$  &  \\
                Unemployed   &  $\unemployment_{i,t}$  &  \\
                Working age rate   & $\nu$   &  $60\%$\\
                Daily productivity per worker  & $\kappa$  &  $\$320.81$ \\
                Fraction of \textbf{infected} individuals that cannot work & $\eta$ & $0.1$ \\
                \hline
                Health index & $\healthindex_i$ & \\
                Economic index & $\economicindex_i$ & \\
                Health welfare weight & $\healthindexweight_i$ & \\
                Social welfare & $\socialwelfare_i$ & $\healthindexweight_i\healthindex_i + (1-\healthindex_i)\economicindex_i$ \\
                \hline
            \end{tabular}
            } %
        \end{center}
        \caption{Model variables and parameters.}
        \label{ext-data-tab:model-parameters}
    \end{small}
\end{table}

\begin{table}[H]
    \begin{small}
        \begin{center}
            {\sffamily %
            \begin{tabular}[c]{lll}
                \hline
                Parameter & Symbol & Value \\
                \hline
                Episode length & $\eplen$ & $540$ \\
                Number of (non-planner) agents & $N$ & $51$ \\
                Time &  $t$  & $1,\ldots, \eplen$ \\
                Agent indices &  $i$  & $1, \ldots, N$\\
                Planner index & $p$ & \\
                Agent policy & $\pi_i$ & \\
                Planner policy & $\pi_p$ & \\
                Model weights &  $\mweight$, $\pweight$  & \\
                State &  $\st$  & \\
                Observation &  $\ob$  & \\
                Action &  $\ac$  & \\
                Reward &  $\rew$  & \\
                Discount factor &  $\df$  & \\
                State-transition, world dynamics &  $\trans$  & \\
                \hline
            \end{tabular}
            } %
        \end{center}
        \caption{Reinforcement learning variables and indices.}
        \label{ext-data-tab:rl-parameters}
    \end{small}
\end{table}